%% file: arxiv.tex
\PassOptionsToPackage{dvipsnames}{xcolor}
\documentclass{bmvc2k}

\title{Test-Time Prototype Adaptation for \\
Open-Vocabulary Semantic Segmentation}

\addauthor{Haozhe Wang}{202481324245@m.scnu.edu.cn}{2}
\addauthor{Jintao Cheng}{jchengau@connect.ust.hk}{1}
\addauthor{Weibin Li}{20228131086@m.scnu.edu.cn}{2}
\addauthor{Xiaoyu Tang\textsuperscript{*}}{tangxy@scnu.edu.cn}{2}

\addinstitution{
  The Hong Kong University of Science and Technology
}
\addinstitution{
  South China Normal University\\
  \textsuperscript{*}Corresponding author
}

\runninghead{Wang, Cheng, Li, Tang}{Test-Time Prototype Adaptation}

\def\ie{\emph{i.e}\bmvaOneDot}

\usepackage[utf8]{inputenc}
\usepackage[T1]{fontenc}
\usepackage{booktabs}
\usepackage{amsfonts}
\usepackage{amsmath}
\usepackage{nicefrac}
\usepackage{microtype}
\usepackage{graphicx}
\usepackage{colortbl}
\usepackage{multirow}
\usepackage{makecell}
\usepackage{url}
\usepackage{algorithm}
\usepackage{algpseudocode}
\usepackage{float}

\begin{document}

\maketitle

\begin{abstract}
\input{sections/arxiv_abstract}
\end{abstract}

\input{sections/arxiv_intro}
\input{sections/arxiv_related}

\input{sections/arxiv_method}
\input{sections/arxiv_experiments}
\input{sections/arxiv_conclusion}

\clearpage
\appendix
\input{sections/arxiv_appendix}

\clearpage
\bibliography{references}

\end{document}

%% file: sections/arxiv_abstract.tex
Open-vocabulary semantic segmentation (OVSS) repurposes a pretrained CLIP encoder for dense prediction without additional labeled supervision. Existing methods improve CLIP's spatial behavior either by redesigning its internal attention or by injecting features from auxiliary vision foundation models—both require access to the host's internal computation and are tailored to its specific forward pass. In this work, we propose \textbf{T}est-time \textbf{P}rototype \textbf{A}daptation (TPA), a training-free plug-in that operates at the output level, leaving the host's forward pass and weights unmodified. By leveraging a lightweight transductive adaptation phase, TPA identifies confident anchor patches from the host's own output predictions on a small pool of unlabeled deployment-domain images, and aggregates their frozen DINO features into per-class prototypes; at inference, a single cosine similarity lookup against this frozen bank provides an auxiliary score fused linearly with the host's logits. TPA composes with five representative OVSS hosts spanning attention-redesign and VFM-injection designs, across three CLIP backbones, eight benchmarks, and multiple internal VFM choices. Under a single set of hyper-parameters and without per-host tuning or parameter updates, TPA consistently improves segmentation accuracy, with as few as approximately 10\% of unlabeled deployment-domain images sufficing for effective bank construction on most benchmarks.

%% file: sections/arxiv_intro.tex
\section{Introduction}
\label{sec:intro}

Open-vocabulary semantic segmentation (OVSS) seeks to assign every pixel a label from an arbitrary set of natural-language categories specified at inference time~\cite{maskclip2022,sclip2024}. The open-vocabulary nature of the task makes CLIP~\cite{clip2021,openclip2022} a natural backbone, as its web-scale image-text pretraining provides strong alignment between visual regions and arbitrary category names. However, CLIP is trained at the image level, and its patch-level features are not directly optimised for dense spatial prediction~\cite{sclip2024,clearclip2024,naclip2024}. Consequently, recent studies have shifted toward methods that improve CLIP's dense behavior at inference without additional labeled supervision. One line of work redesigns CLIP's internal attention to enhance spatial coherence~\cite{maskclip2022,sclip2024,clearclip2024,naclip2024,gem2024}. Another injects attention or spatial priors from auxiliary vision foundation models (VFMs) such as DINO~\cite{dino2021,dinov22023} or SAM~\cite{sam2023} into CLIP's pipeline~\cite{proxyclip2024,trident2024}. Recent plug-in extensions further refine the host at inference, either by modulating its intermediate attention via output feedback~\cite{fsa2025} or by gradient-based entropy minimization~\cite{mlmp2025}.

More broadly, dense visual perception in robotics often relies on task-specific spatial structure, as illustrated by specialized instance segmentation for road-surface hazards~\cite{cheng2023pothole}, motion-aware representations~\cite{cheng2024mf}, and multi-view feature fusion~\cite{cheng2024mv}. These examples highlight the value of incorporating complementary evidence beyond a single local prediction, while leaving open the question of how to obtain such cues for open-vocabulary segmentation without task-specific training or redesigning the host model.

\begin{figure*}[t]
  \centering
  \includegraphics[width=\textwidth]{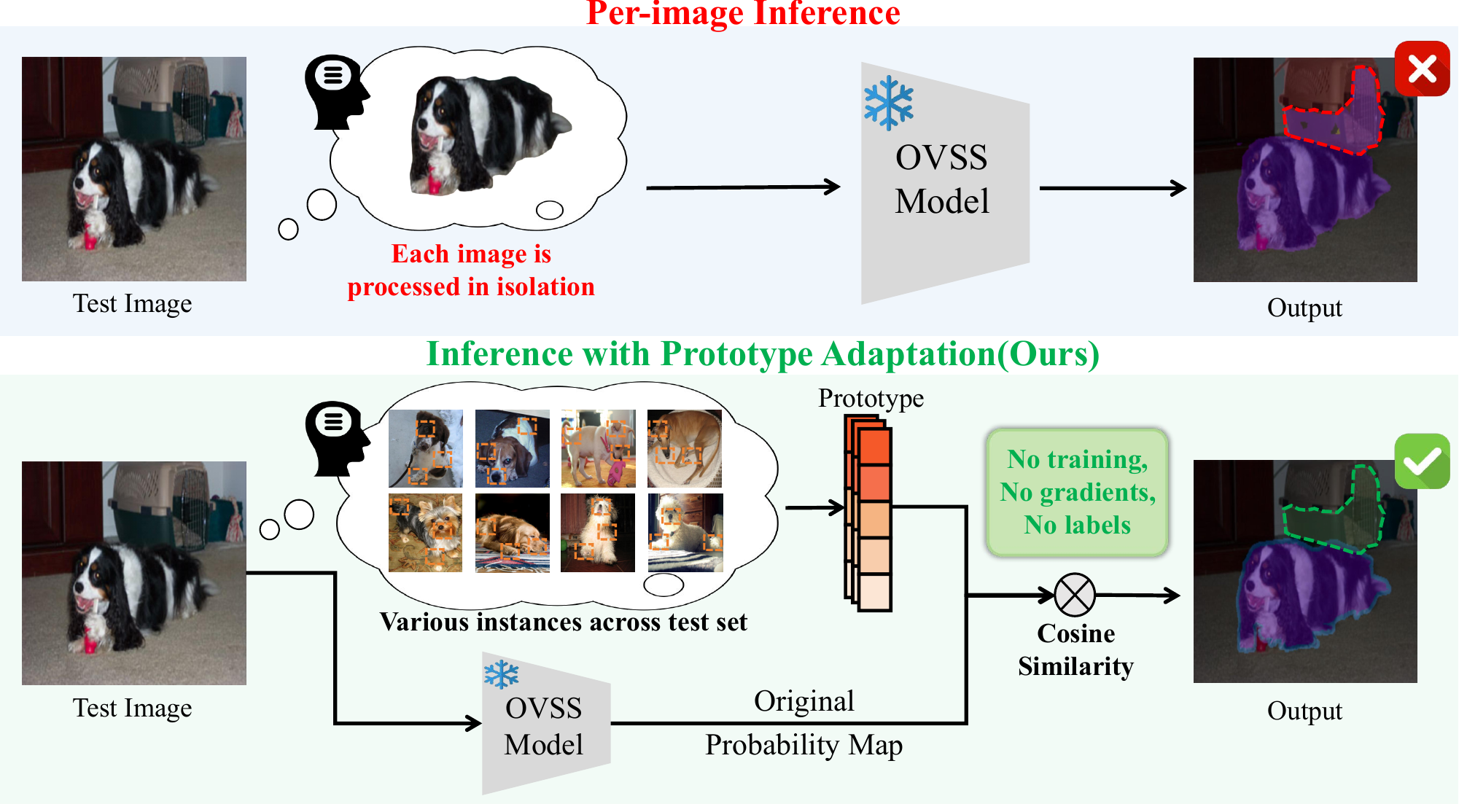}
  \caption{Overview of TPA. A lightweight adaptation phase constructs a per-class prototype bank from host-confident patches encoded by a frozen DINO model, requiring only a small pool of unlabeled deployment-domain images. At inference, each pixel receives an auxiliary class score via a single cosine similarity lookup against the bank, linearly fused with the host's original logits. The host's forward pass and weights are never modified.}
  \label{fig:teaser}
\end{figure*}

While highly effective, these approaches share a common reliance on intervening within the host's internal computation. Attention redesign methods alter CLIP's self-attention structure for a particular architecture~\cite{sclip2024,clearclip2024,naclip2024}. VFM-injection and feedback-based plug-ins integrate auxiliary features or modulate intermediate attention maps at specific network layers~\cite{proxyclip2024,trident2024,fsa2025}. Consequently, these designs are inherently coupled with the internal structure of a specific host. Applying them to a new model requires bespoke engineering of its forward pass.

However, the host's final output predictions already carry informative class-level signals. In typical deployment scenarios, instances of the same semantic category recur across images, exhibiting different viewpoints and contexts. A dog, for instance, appears in canonical frontal poses in some images but in unusual angles in others. The host's output predictions are naturally highly confident on many of these canonical instances. Aggregating the visual features of such confident patches into a \textit{class prototype} yields a stable representation robust to per-instance variability. Extracting stable cross-image priors from a lightweight adaptation pool prior to inference naturally bypasses architectural constraints. This strategy corrects uncertain predictions purely at the output level, yielding substantial performance gains without requiring complex internal modifications.

Building upon this observation, we propose \textbf{T}est-time \textbf{P}rototype \textbf{A}daptation (TPA), a training-free plug-in for OVSS (Fig.~\ref{fig:teaser}). Following a transductive test-time adaptation protocol~\cite{t3a2021,lame2022}, TPA leverages a small pool of unlabeled deployment images to construct a frozen prototype bank. To prevent high-confidence hallucinations from polluting the prototypes, TPA employs a robust filtering mechanism to isolate reliable anchor patches. Leveraging the object-centric spatial clustering of self-supervised DINO representations~\cite{dino2021,dinov22023,clipdinoiser2023}, it extracts features at these anchor locations, aggregating them into per-class prototypes. At inference, each pixel receives an auxiliary score via a cosine similarity lookup against the bank, linearly fused with the host's original logits.

To validate TPA, we apply it to five representative OVSS hosts (SCLIP~\cite{sclip2024}, NACLIP~\cite{naclip2024}, ClearCLIP~\cite{clearclip2024}, ProxyCLIP~\cite{proxyclip2024}, and Trident~\cite{trident2024}). The evaluation covers three CLIP backbones, eight benchmarks, and multiple internal VFM choices. TPA consistently improves segmentation accuracy under a single set of hyper-parameters. Notably, operating strictly at the output level allows TPA to seamlessly compose with internal plug-ins, yielding further cumulative gains when stacked with methods like MLMP~\cite{mlmp2025} or FSA~\cite{fsa2025}. Furthermore, requiring as few as 10\% of the deployment images for effective bank construction, TPA introduces minimal transductive overhead alongside a negligible inference-time cosine lookup.

\paragraph{Contributions.}
\begin{itemize}
  \item We propose TPA, a training-free, host-agnostic plug-in for OVSS. Operating purely at the output level, it avoids modifying the host's forward pass or weights, ensuring seamless portability across diverse architectures.
  \item TPA extracts shared visual patterns from a lightweight unlabeled deployment pool into DINO-space prototypes, employing robust anchor filtering to mitigate confirmation bias.
  \item Across five hosts, three backbones, and eight benchmarks, TPA consistently improves accuracy and exhibits strong complementarity when stacked with internal adaptation methods.
  \item Comprehensive analyses demonstrate TPA's efficiency, requiring only a fraction of deployment images for prototype construction and a simple cosine lookup at inference.
\end{itemize}

%% file: sections/arxiv_related.tex
\section{Related Work}
\label{sec:related}

\paragraph{Open-vocabulary semantic segmentation.}
CLIP's large-scale image-text pretraining~\cite{clip2021,openclip2022} provides strong visual-text alignment for open-vocabulary recognition. Repurposing CLIP for dense prediction, however, is challenging because its global-level pretraining leads to noisy patch-level localization~\cite{maskclip2022,sclip2024}. Recent training-free methods address this by modulating CLIP's internal computation. Along the first axis, attention redesign methods alter CLIP's self-attention to improve spatial coherence. MaskCLIP~\cite{maskclip2022} retains only the value projection; SCLIP~\cite{sclip2024} replaces query-key attention with symmetric self-attentions; NACLIP~\cite{naclip2024} injects a locality prior; GEM~\cite{gem2024} uses key-key attention as a grounding signal; and ClearCLIP~\cite{clearclip2024} removes the residual connection in the final block. Along the second axis, VFM-injection methods integrate spatial priors from stronger vision encoders into CLIP's pipeline. ProxyCLIP~\cite{proxyclip2024} replaces CLIP's attention with that of DINO~\cite{dino2021,dinov22023}, and Trident~\cite{trident2024} further combines DINO and SAM~\cite{sam2023} for mask-level refinement. Along the third axis, plug-in extensions refine the host at inference time. FSA~\cite{fsa2025} modulates each host's intermediate attention via a feedback loop from its own output predictions, and MLMP~\cite{mlmp2025} performs gradient-based entropy minimization at each test image. However, all of these approaches intervene within the host's forward pass, leaving the class-level signals encoded in the host's output predictions unexploited as a source for dense-prediction refinement.

\paragraph{Test-time adaptation and prototype-based inference.}
Test-time adaptation (TTA) improves model performance on unlabeled deployment data without retraining. Gradient-based methods such as TENT~\cite{tent2021}, EATA~\cite{eata2022}, and CoTTA~\cite{cotta2022} update model parameters by minimizing entropy or related objectives. Beyond classification, efficient test-time guidance has also been explored for multimodal visual place recognition as an alternative to conventional fine-tuning~\cite{cheng2025scale}, reflecting a broader interest in adapting pretrained models at deployment with limited additional optimization. Parameter-free methods go further by operating without gradient access. T3A~\cite{t3a2021} builds class-level templates from the model's own high-confidence predictions at test time and classifies subsequent inputs by nearest-prototype assignment. LAME~\cite{lame2022} refines output posteriors via a Laplacian regularizer without parameter updates. These prototype-based TTA methods share the core mechanism with our approach, but are designed for image-level classification and do not address dense prediction or vision-language alignment. In this work, we bring the prototype-based inference principle to open-vocabulary dense prediction in a fully gradient-free, training-free manner. TPA constructs per-class prototypes from confident host predictions over a small unlabeled deployment-domain pool in DINO feature space, and fuses the resulting cosine similarity signal with the host's original logits. The entire process is gradient-free, label-free, and external to the host's forward pass, making TPA applicable to any OVSS host regardless of its internal architecture or internal VFM choice.

%% file: sections/arxiv_method.tex
\section{Method}
\label{sec:method}

\begin{figure*}[t]
    \centering
    \includegraphics[width=\textwidth]{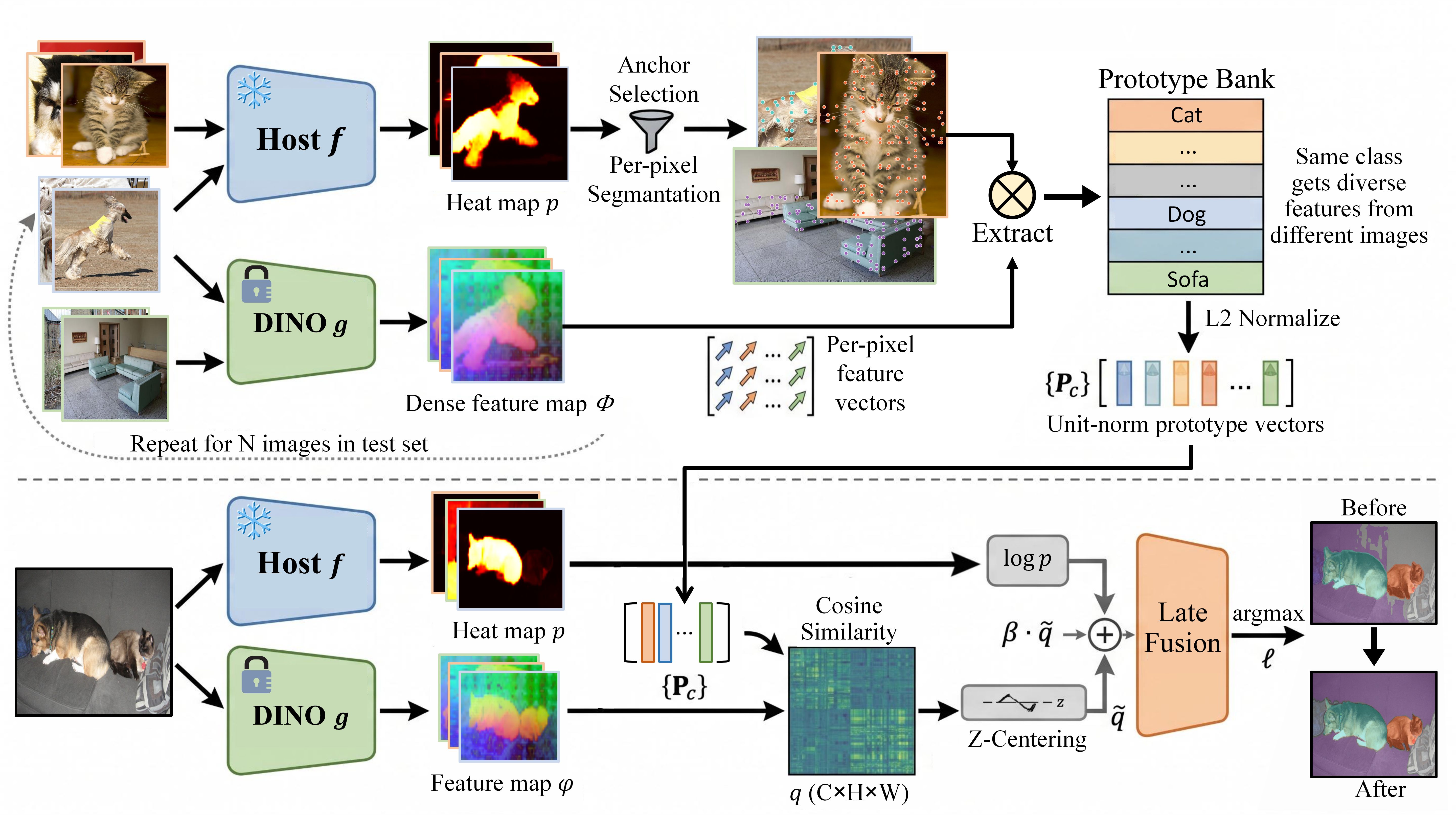}
    \caption{Overview of TPA. \emph{Adaptation phase:} the host model~$f$ produces per-pixel class probabilities from a small unlabeled deployment-domain pool; high-confidence anchor patches identified from $\mathbf{p}$ contribute their DINO features to a per-class running statistic. \emph{Inference:} the statistic yields unit-norm class prototypes that define an auxiliary dense classifier, whose cosine scores are linearly fused with the host's log-probabilities. Both $f$ and $g$ remain frozen throughout.}
    \label{fig:overview}
\end{figure*}

In this section, we present the TPA framework (Fig.~\ref{fig:overview}). We first establish the problem formulation and the roles of the two foundation models (Sec.~\ref{sec:method-setup}). We then describe anchor selection and prototype construction during the adaptation phase (Sec.~\ref{sec:method-pass1}) and the prototype-based auxiliary classifier and late fusion at inference (Sec.~\ref{sec:method-pass2}). Finally, we summarise the key properties and computational cost (Sec.~\ref{sec:method-properties}).

\subsection{Problem formulation}
\label{sec:method-setup}
We consider a frozen OVSS host model $f$. Given an image $I$ and a fixed set of text queries $\mathcal{C}=\{t_1,\dots,t_C\}$, the host produces a dense probability tensor
\begin{equation}
  \mathbf{p} \;=\; f(I;\mathcal{C}) \;\in\; (\Delta^{C-1})^{H\times W},
  \quad
  \mathbf{p}(c,i,j)=\Pr\!\bigl(\text{pixel }(i,j)\text{ belongs to class }c\,\bigm|\,I,\mathcal{C}\bigr),
  \label{eq:host-output}
\end{equation}
where $\Delta^{C-1}$ denotes the $(C\!-\!1)$-simplex. The internal mechanism of $f$---whether attention redesign, VFM injection, or any other approach---is immaterial; the only interface we require is the dense output tensor $\mathbf{p}$. Because $f$ is pretrained with image-level objectives, its patch-level predictions are noisy: confident predictions and uncertain predictions coexist in every output map. The key insight of TPA is that reliable predictions already present in $\mathbf{p}$ can be leveraged to construct stable class-level references, which in turn correct the uncertain ones.

To realise this idea, we use a frozen DINO encoder~\cite{dino2021} as an external feature extractor $g$, producing a dense feature map
\begin{equation}
  \boldsymbol{\phi} \;=\; g(I) \;\in\; \mathbb{R}^{D\times H\times W},
\end{equation}
bilinearly upsampled to the pixel grid of $\mathbf{p}$. DINO features are well-suited for this role: self-supervised training endows them with strong spatial coherence and object-centric grouping~\cite{dino2021,clipdinoiser2023}, providing a stable space for prototype aggregation. Although $g$ operates at a lower spatial resolution (e.g., patch size 14), fusing its semantic region guidance in the logit space ensures that the high-frequency spatial details of the host's original prediction $\mathbf{p}$ are preserved. Crucially, $g$ is never inserted into the forward pass of $f$; it operates as a separate, read-only encoder. We never update the weights of $f$ or $g$ and do not use any ground-truth labels.

\subsection{Anchor selection and prototype construction}
\label{sec:method-pass1}

\noindent\textbf{Anchor selection.}~~
Although $\mathbf{p}$ is globally noisy, many pixels in every image are predicted with high confidence and are unlikely to be erroneous. We exploit this by identifying, for each class $c$, the pixels at which the host's prediction is both argmax-consistent and above a confidence floor. A pixel $(i,j)$ is declared a class-$c$ \emph{anchor} when
\begin{equation}
  \arg\max_{c'}\,\mathbf{p}(c',i,j)=c
  \quad\text{and}\quad
  \mathbf{p}(c,i,j)>\tau,
  \label{eq:anchor-rule}
\end{equation}
where $\tau = 2/C$ is a dataset-adaptive threshold derived from the number of target classes $C$. This requires each anchor to carry at least twice the expected probability under the null (uniform) distribution $1/C$, filtering out near-random predictions while scaling naturally with the label space. Crucially, this threshold, working in tandem with the strict argmax condition, forms the first line of defense against confirmation bias and high-confidence hallucinations on novel classes. As empirically validated by the qualitative visualizations in Sec.~\ref{sec:exp-qualitative} (Fig.~\ref{fig:anchor-vis}), this dual-condition filtering effectively isolates spatially coherent, class-specific semantic regions, ensuring high anchor quality without requiring manual tuning.

Let $\mathcal{A}^c$ denote the set of pixels satisfying Eq.~\eqref{eq:anchor-rule}. We retain all of $\mathcal{A}^c$ provided $|\mathcal{A}^c|\geq K_{\min}$; for classes whose qualifying pixel count falls below $K_{\min}$, we construct no prototype and leave the host prediction unchanged. Skipping under-represented classes is preferable to forcing low-quality anchors from below-threshold pixels, acting as an additional safety gate against spurious predictions.

\noindent\textbf{Prototype construction.}~~
Each anchor contributes its DINO feature to a per-class running sufficient statistic maintained across the adaptation pool $\{I_1,\dots,I_M\}$:
\begin{equation}
  \mathbf{s}_c \;=\; \sum_{n=1}^{M}\;\sum_{(i,j)\in\mathcal{A}_n^c}\!\boldsymbol{\phi}_n(i,j),
  \qquad
  N_c \;=\; \sum_{n=1}^{M}\bigl\lvert\mathcal{A}_n^c\bigr\rvert,
  \label{eq:stats}
\end{equation}
requiring $\mathcal{O}(CD)$ memory in total. Beyond simple aggregation, the cross-image running sum serves to further dilute any sporadic false positives that might have survived the pixel-level filtering. At the end of the adaptation phase, we form the unit-norm prototype for each class that received at least one anchor:
\begin{equation}
  \mathbf{P}_c \;=\; \begin{cases}
    \mathbf{s}_c/\lVert\mathbf{s}_c\rVert_2, & N_c>0,\\[2pt]
    \mathbf{0}, & N_c=0,
  \end{cases}
  \qquad
  \mathcal{C}^{\star}\!=\!\{c:\,N_c>0\}.
  \label{eq:prototype}
\end{equation}
Classes not represented in $\mathcal{C}^{\star}$ will be unaffected by the auxiliary classifier described next. Because Eq.~\eqref{eq:stats} is a running sum, new images can be folded into the statistic incrementally without revisiting previous ones.

\subsection{Prototype-based auxiliary classifier and late fusion}
\label{sec:method-pass2}

\noindent\textbf{Cosine auxiliary classifier.}~~
At inference, we compare each pixel's L2-normalised DINO feature $\hat{\boldsymbol{\phi}}(i,j)$ against every class prototype to obtain an auxiliary logit:
\begin{equation}
  \mathbf{q}(c,i,j)\;=\;\mathbf{P}_c^{\top}\hat{\boldsymbol{\phi}}(i,j)\cdot\mathbb{1}\!\left[c\in\mathcal{C}^{\star}\right].
  \label{eq:dino-logit}
\end{equation}
Due to the anisotropic nature of self-supervised feature spaces like DINO (often referred to as representation collapse), all feature vectors tend to lie in a narrow cone. Consequently, the raw cosine similarities $\mathbf{q}$ carry a large, positive per-pixel mean offset. If uncorrected, this global offset would overwhelm the relative class differences when fused with the host's logits. We remove this offset by centring $\mathbf{q}$ across the available prototypes,
\begin{equation}
  \bar{\mathbf{q}}(i,j)\;=\;\frac{1}{\lvert\mathcal{C}^{\star}\rvert}\sum_{c\in\mathcal{C}^{\star}}\!\mathbf{q}(c,i,j),
  \qquad
  \tilde{\mathbf{q}}(c,i,j)\;=\;\bigl(\mathbf{q}(c,i,j)-\bar{\mathbf{q}}(i,j)\bigr)\cdot\mathbb{1}\!\left[c\in\mathcal{C}^{\star}\right],
  \label{eq:zcenter}
\end{equation}
so that $\tilde{\mathbf{q}}$ encodes only the \emph{relative} preference of the pixel among prototype-endowed classes, and vanishes at pixels where all prototypes score equally (\ie where the prototype bank provides no discriminative evidence).

\noindent\textbf{Late fusion.}~~
The auxiliary logit $\tilde{\mathbf{q}}$ is combined with the host's output in log-space, treating the two sources as independent evidence:
\begin{equation}
  \boxed{\;\ell(c,i,j)\;=\;\log\mathbf{p}(c,i,j)\;+\;\beta\,\tilde{\mathbf{q}}(c,i,j),\qquad\beta=(1-\alpha)\,\lambda,\;}
  \label{eq:fusion}
\end{equation}
where $\alpha\in[0,1]$ is a fusion coefficient and $\lambda$ is a temperature that rescales $\tilde{\mathbf{q}}$ to the dynamic range of $\log\mathbf{p}$ (fixed to $\lambda=5$ throughout). For classes outside $\mathcal{C}^{\star}$, $\tilde{\mathbf{q}}$ vanishes and Eq.~\eqref{eq:fusion} reduces to the host's original log-probability. While zero-filling for unrepresented classes might theoretically induce a slight calibration shift in softmax assignment, the fixed coefficient $\alpha$ ensures the host's output remains dominant. Moreover, as demonstrated in Sec.~\ref{sec:exp-efficiency}, virtually all dataset classes rapidly accumulate sufficient anchors to enter $\mathcal{C}^{\star}$ within the first few images, rendering this edge case negligible in practice. The predicted label is $\hat{y}(i,j)=\arg\max_c\ell(c,i,j)$.

\subsection{Properties and cost}
\label{sec:method-properties}
Three properties of TPA bear directly on the experimental design of Section~\ref{sec:experiments}. First, it is \emph{host-agnostic}: only the dense tensor $\mathbf{p}$ of Eq.~\eqref{eq:host-output} is read from $f$; no internal attention, hidden state, or gradient is accessed. The same procedure therefore attaches to any of the OVSS hosts in Section~\ref{sec:related} without modification. Second, it is \emph{gradient- and label-free}: no parameters are updated and no ground-truth label enters Eqs.~\eqref{eq:anchor-rule}--\eqref{eq:fusion}. Third, the compute overhead amounts to one forward pass of $f$ and $g$ per adaptation-pool image---performed once before inference---plus $\mathcal{O}(CD)$ storage for the statistics in Eq.~\eqref{eq:stats}; at inference, the only additional cost per image is a single cosine similarity lookup of dimension $D$ per pixel against the frozen prototype bank. The anchor threshold $\tau = 2/C$ is determined analytically from the label-space size and requires no tuning. The remaining parameters --- $K_{\min}$ (minimum anchor count per class, fixed to 5), $\alpha$ (fusion weight), and $\lambda$ (logit temperature) --- are scalar and shared across all classes, datasets, and hosts.

%% file: sections/arxiv_experiments.tex
\section{Experiments}
\label{sec:experiments}

\subsection{Experimental set-up}
\label{sec:exp-setup}

\paragraph{Hosts and backbones.}
We apply TPA to a set of training-free OVSS hosts spanning the design space reviewed in Section~\ref{sec:related}: SCLIP~\cite{sclip2024} (symmetric self-attention), ClearCLIP~\cite{clearclip2024} (residual removal), NACLIP~\cite{naclip2024} (locality-prior attention), ProxyCLIP~\cite{proxyclip2024} (DINO-proxy attention), and Trident~\cite{trident2024} (DINO proxy with SAM mask refinement). We evaluate ten host$\times$backbone configurations across ViT-B/16, ViT-L/14, and ViT-H/14.

\paragraph{Datasets.}
Following the standard OVSS evaluation protocol~\cite{naclip2024,proxyclip2024,clearclip2024}, we report mean intersection-over-union (mIoU) on eight benchmarks: PASCAL~VOC (21 and 20 classes)~\cite{pascalvoc2010}, PASCAL~Context (59 and 60 classes)~\cite{pcontext2014}, ADE20K~\cite{ade20k2019}, COCO-Object~\cite{coco2014}, COCO-Stuff~\cite{cocostuff2018}, and Cityscapes~\cite{cityscapes2016}.

\paragraph{Our procedure.}
Unless stated otherwise, we use a single set of hyper-parameters across \emph{all} host$\times$backbone$\times$dataset configurations: minimum anchor count $K_{\min}{=}5$, fusion coefficient $\alpha{=}0.5$, and temperature $\lambda{=}5$. The anchor-selection threshold is set as $\tau = 2/C$, where $C$ is the benchmark's class count; this is a fixed formula requiring no tuning (see Section~\ref{sec:method-pass1}). The auxiliary encoder $g$ is DINOv2 ViT-B/14~\cite{dinov22023}. Following the transductive test-time adaptation protocol~\cite{t3a2021,lame2022}, the adaptation pool for each benchmark is drawn from the same test domain with labels withheld; pool size sensitivity is examined in Sec.~\ref{sec:exp-ablation}.

\paragraph{Baseline reporting.}
\textsc{Base} values are taken from the reproduction of~\cite{fsa2025} whenever the host is covered there (SCLIP, ClearCLIP, ProxyCLIP); for NACLIP and Trident we use our own reproduction, which agrees with the reported values to within $\pm1.5$\,mIoU. \textsc{+FSA}~\cite{fsa2025} rows are taken verbatim from~\cite{fsa2025}; \textsc{+ResCLIP}~\cite{resclip2025} rows for SCLIP~B/16 and ClearCLIP~B/16 are paper values; the NACLIP~B/16 \textsc{+ResCLIP} row is from our own evaluation following the ResCLIP source code, as~\cite{resclip2025} does not report NACLIP. The composition table (\S\ref{sec:exp-composition}) uses the MLMP evaluation protocol as its NACLIP~L/14 baseline; values in that block are under this unified protocol.

\subsection{Main results}
\label{sec:exp-main}

\begin{table*}[t]
  \caption{\textbf{Main results} (ViT-B/16 backbone; full results in the Appendix). \textsc{Base} rows report the unmodified host; \textsc{+FSA} and \textsc{+ResCLIP} rows are from~\cite{fsa2025,resclip2025} where available. $\Delta$ is computed against the \textsc{Base} row. \textcolor{BrickRed}{Red}: FSA / ResCLIP delta; \textcolor{ForestGreen}{green}: ours. ($^{\ast}$) Evaluated following ResCLIP source code.}
  \label{tab:main}
  \centering
  \footnotesize
  \resizebox{\linewidth}{!}{%
  \begin{tabular}{l c c c c c c c c c}
    \toprule
    \textbf{Host / Method} &
    VOC & Ctx & Obj & VOC20 & Ctx59 & Stuff & City & ADE & \textbf{Avg.} \\
    \midrule
    SCLIP~\cite{sclip2024}
      & 59.1 & 30.4 & 30.5 & 80.4 & 34.2 & 22.4 & 32.2 & 16.1 & 38.2 \\
    \quad +FSA~\cite{fsa2025}
      & 61.5 & 33.3 & 33.9 & 82.8 & 36.8 & 24.4 & 34.7 & 17.5 & 40.6\,\textcolor{BrickRed}{\scriptsize(+2.4)} \\
    \quad +ResCLIP~\cite{resclip2025}
      & 60.7 & 32.9 & 34.3 & 84.6 & 35.8 & 23.9 & 34.4 & 17.6 & 40.5\,\textcolor{BrickRed}{\scriptsize(+2.3)} \\
    \rowcolor{gray!10}
    \quad +Ours
      & 61.1 & 34.4 & 32.7 & 84.4 & 39.0 & 26.0 & 38.2 & 19.1 & \textbf{41.9}\,\textcolor{ForestGreen}{\scriptsize(+3.7)} \\
    \midrule
    ClearCLIP~\cite{clearclip2024}
      & 51.8 & 32.6 & 33.0 & 80.9 & 35.9 & 23.9 & 30.0 & 16.7 & 38.1 \\
    \quad +FSA~\cite{fsa2025}
      & 53.0 & 33.6 & 33.2 & 81.3 & 33.8 & 24.3 & 30.8 & 17.4 & 38.8\,\textcolor{BrickRed}{\scriptsize(+0.7)} \\
    \quad +ResCLIP~\cite{resclip2025}
      & 59.0 & 32.9 & 34.0 & 87.1 & 36.4 & 24.3 & 34.5 & 17.8 & 40.8\,\textcolor{BrickRed}{\scriptsize(+2.7)} \\
    \rowcolor{gray!10}
    \quad +Ours
      & 53.9 & 37.3 & 35.8 & 83.7 & 39.7 & 26.0 & 34.8 & 18.4 & \textbf{41.2}\,\textcolor{ForestGreen}{\scriptsize(+3.1)} \\
    \midrule
    NACLIP~\cite{naclip2024}
      & 64.1 & 37.7 & 36.2 & 83.0 & 40.0 & 25.7 & 38.3 & 19.1 & 43.0 \\
    \quad +ResCLIP~\cite{resclip2025}$^{\ast}$
      & 65.0 & 38.2 & 37.4 & 85.0 & 41.0 & 26.5 & 39.0 & 19.8 & 44.0\,\textcolor{BrickRed}{\scriptsize(+1.0)} \\
    \rowcolor{gray!10}
    \quad +Ours
      & 66.5 & 40.8 & 38.2 & 86.3 & 43.5 & 28.3 & 42.8 & 22.0 & \textbf{46.1}\,\textcolor{ForestGreen}{\scriptsize(+3.1)} \\
    \midrule
    ProxyCLIP~\cite{proxyclip2024}
      & 61.3 & 35.3 & 37.5 & 80.3 & 39.1 & 26.5 & 38.1 & 20.2 & 42.3 \\
    \quad +FSA~\cite{fsa2025}
      & 63.7 & 36.1 & 38.0 & 82.3 & 39.9 & 27.0 & 38.8 & 20.5 & 43.3\,\textcolor{BrickRed}{\scriptsize(+1.0)} \\
    \rowcolor{gray!10}
    \quad +Ours
      & 62.6 & 37.8 & 38.7 & 81.7 & 41.5 & 28.7 & 41.2 & 22.4 & \textbf{44.3}\,\textcolor{ForestGreen}{\scriptsize(+2.0)} \\
    \midrule
    Trident~\cite{trident2024}
      & 67.1 & 42.0 & 41.1 & 84.5 & 44.2 & 28.3 & 42.9 & 21.9 & 46.5 \\
    \rowcolor{gray!10}
    \quad +Ours
      & 67.4 & 44.2 & 41.8 & 86.4 & 47.0 & 30.5 & 45.7 & 24.3 & \textbf{48.4}\,\textcolor{ForestGreen}{\scriptsize(+1.9)} \\
    \bottomrule
  \end{tabular}}
\end{table*}

\noindent Table~\ref{tab:main} reports mIoU on ViT-B/16 for five representative hosts; complete results across all backbones in the Appendix.

\paragraph{Consistent improvement.}
TPA yields a positive $\Delta$\,mIoU on every B/16 host: SCLIP~($+3.7$), ClearCLIP~($+3.1$), NACLIP~($+3.1$), ProxyCLIP~($+2.0$), and Trident~($+1.9$). The pattern holds across all backbones (Appendix), with a mean gain of $+2.8$\,mIoU over the full ten-configuration evaluation. Gains are largest on attention-redesign hosts where CLIP's per-patch localisation is weakest; ProxyCLIP and Trident, which already inject per-image DINO or SAM features, show smaller but consistent improvements.

\paragraph{Comparison with per-image plug-ins.}
On the three B/16 hosts covered by FSA~\cite{fsa2025}, TPA outperforms it on all three: SCLIP~($+3.7$ vs.\ $+2.4$), ClearCLIP~($+3.1$ vs.\ $+0.7$), and ProxyCLIP~($+2.0$ vs.\ $+1.0$). FSA narrows the gap on ProxyCLIP~L/14 and H/14 (Appendix) by directly modulating the DINO-proxy attention that ProxyCLIP itself constructs, an architecture-specific entry point effective for ProxyCLIP but non-transferable to other hosts. This last point explains the coverage pattern in Table~\ref{tab:main}. FSA hooks into CLIP's QKV tensors, so it cannot be applied to NACLIP, whose locality-prior replaces standard self-attention with a modified locality-constrained variant; it is also not evaluated on Trident in~\cite{fsa2025}. ResCLIP~\cite{resclip2025} is similarly bounded by its residual-stream assumptions, and MLMP~\cite{mlmp2025} by dataset-specific prompt files; neither~\cite{resclip2025} nor~\cite{fsa2025} reports results for NACLIP or Trident. Reading only the post-softmax output tensor~$\mathbf{p}$, defined uniformly across all hosts, TPA applies to all five B/16 hosts without modification, including NACLIP ($+3.1$) and Trident ($+1.9$), which prior per-image methods cannot reach.

\subsection{Generalisation across auxiliary VFMs}
\label{sec:exp-vfm}

The bank extractor $g$ is fixed to DINOv2 ViT-B/14 throughout, independent of the VFM the host uses internally. To confirm that the benefit is not an artefact of matching the host's internal VFM, we apply TPA (unchanged) on top of ProxyCLIP when its internal proxy is switched from DINOv2 to SAM~\cite{sam2023} ViT-B/16 and MAE~\cite{mae2022} ViT-B/16. Table~\ref{tab:vfm} shows ViT-B/16; results for ViT-L/14 and ViT-H/14 (27 groups in total) are in the Appendix.

\begin{table*}[t]
  \caption{\textbf{Generalisation across ProxyCLIP's internal VFM} (ViT-B/16 backbone). The TPA bank extractor (DINOv2) is held fixed; only the host's internal proxy varies. Proxy and \textsc{+FSA} rows from~\cite{fsa2025}.}
  \label{tab:vfm}
  \centering
  \footnotesize
  \resizebox{\linewidth}{!}{%
  \begin{tabular}{l l c c c c c c c c c}
    \toprule
    \textbf{VFM} & \textbf{Method} &
    VOC & Ctx & Obj & VOC20 & Ctx59 & Stuff & City & ADE & \textbf{Avg.} \\
    \midrule
    \multirow{3}{*}{\makecell{SAM\\ViT-B/16}}
      & Proxy
                   & 59.3 & 33.6 & 35.4 & 80.4 & 37.0 & 25.0 & 37.0 & 19.1 & 40.8 \\
    & \quad+FSA~\cite{fsa2025}
                  & 60.7 & 34.0 & 35.8 & 81.8 & 37.4 & 25.2 & 37.9 & 19.3 & 41.5\,\textcolor{BrickRed}{\scriptsize(+0.7)} \\
    & \cellcolor{gray!10}\quad+Ours & \cellcolor{gray!10}61.3 & \cellcolor{gray!10}34.9 & \cellcolor{gray!10}38.7 & \cellcolor{gray!10}83.2 & \cellcolor{gray!10}38.8 & \cellcolor{gray!10}26.9 & \cellcolor{gray!10}42.1 & \cellcolor{gray!10}20.3 & \cellcolor{gray!10}\textbf{43.3}\,\textcolor{ForestGreen}{\scriptsize(+2.5)} \\
    \cmidrule(lr){1-11}
    \multirow{3}{*}{\makecell{MAE\\ViT-B/16}}
      & Proxy
                   & 52.2 & 30.4 & 30.8 & 76.3 & 33.5 & 23.1 & 30.1 & 17.1 & 36.7 \\
    & \quad+FSA~\cite{fsa2025}
                  & 54.3 & 30.9 & 31.3 & 78.1 & 33.9 & 23.4 & 33.6 & 17.5 & 37.9\,\textcolor{BrickRed}{\scriptsize(+1.2)} \\
    & \cellcolor{gray!10}\quad+Ours & \cellcolor{gray!10}56.9 & \cellcolor{gray!10}33.4 & \cellcolor{gray!10}35.6 & \cellcolor{gray!10}81.4 & \cellcolor{gray!10}37.6 & \cellcolor{gray!10}25.6 & \cellcolor{gray!10}37.1 & \cellcolor{gray!10}17.9 & \cellcolor{gray!10}\textbf{40.7}\,\textcolor{ForestGreen}{\scriptsize(+4.0)} \\
    \cmidrule(lr){1-11}
    \multirow{3}{*}{\makecell{DINOv2\\ViT-B/14}}
      & Proxy
                   & 58.6 & 33.8 & 37.0 & 83.0 & 37.2 & 25.4 & 33.9 & 19.7 & 41.1 \\
    & \quad+FSA~\cite{fsa2025}
                  & 59.2 & 33.9 & 37.4 & 84.0 & 37.5 & 25.5 & 34.4 & 19.7 & 41.4\,\textcolor{BrickRed}{\scriptsize(+0.3)} \\
    & \cellcolor{gray!10}\quad+Ours & \cellcolor{gray!10}60.3 & \cellcolor{gray!10}34.9 & \cellcolor{gray!10}38.5 & \cellcolor{gray!10}84.8 & \cellcolor{gray!10}38.7 & \cellcolor{gray!10}26.7 & \cellcolor{gray!10}35.9 & \cellcolor{gray!10}20.9 & \cellcolor{gray!10}\textbf{42.6}\,\textcolor{ForestGreen}{\scriptsize(+1.5)} \\
    \bottomrule
  \end{tabular}}
\end{table*}

TPA (DINOv2 bank, fixed) yields a positive average gain across every ProxyCLIP VFM configuration and benchmark on ViT-B/16. The margin is largest when the host uses MAE as its proxy ($+4.0$ vs.\ FSA's $+1.2$), where weaker per-image MAE localisation leaves more room for cross-image DINO prototypes. The same pattern holds across all 27 groups spanning three backbones (Appendix), confirming that TPA's benefit is orthogonal to the host's internal VFM choice.

\subsection{Ablation studies}
\label{sec:exp-ablation}

\noindent\textbf{Component ablation.}~~
We ablate the four core design choices of TPA on NACLIP~B/16, reporting $\Delta$\,mIoU on three diverse benchmarks (VOC21, Context59, ADE20K).

\begin{table}[t]
  \begin{minipage}[t]{0.60\linewidth}
    \centering
    \footnotesize
    \resizebox{\linewidth}{!}{%
    \begin{tabular}{l c c c c}
      \toprule
      \textbf{Variant} & \textbf{VOC21} & \textbf{Ctx59} & \textbf{ADE} & \textbf{Avg.} \\
      \midrule
      Per-image prototypes (no pool) & +0.7 & +1.0 & +0.3 & +0.7 \\
      CLIP patch features (replace DINO) & +0.8 & +1.2 & +0.4 & +0.8 \\
      Random anchor selection (same $K$) & +0.9 & +1.4 & +0.4 & +0.9 \\
      No centring (raw cosine fusion) & +1.3 & +1.9 & +0.6 & +1.3 \\
      \rowcolor{gray!10}
      \textbf{TPA (full)} & \textbf{+2.4} & \textbf{+3.5} & \textbf{+2.9} & \textbf{+2.9} \\
      \bottomrule
    \end{tabular}}
    \caption{\textbf{Component ablation on NACLIP~B/16.} Each row removes or replaces one design choice while keeping others at default.}
    \label{tab:ablation}
  \end{minipage}%
  \hfill%
  \begin{minipage}[t]{0.37\linewidth}
    \centering
    \footnotesize
    \resizebox{\linewidth}{!}{%
    \begin{tabular}{c c c c c}
      \toprule
      $k$ & \textbf{VOC21} & \textbf{Ctx59} & \textbf{ADE} & \textbf{Avg.} \\
      \midrule
      1.0 & +2.28 & +3.42 & +2.77 & +2.82 \\
      1.5 & +2.38 & +3.49 & +2.86 & +2.91 \\
      \rowcolor{gray!10}
      \textbf{2.0}$^\dagger$ & \textbf{+2.40} & \textbf{+3.50} & \textbf{+2.92} & \textbf{+2.94} \\
      2.5 & +2.39 & +3.47 & +2.86 & +2.91 \\
      3.0 & +2.27 & +3.36 & +2.72 & +2.78 \\
      \bottomrule
    \end{tabular}}
    \caption{\textbf{Sensitivity to $k$ in $\tau{=}k/C$.} ($\dagger$) marks the default $k{=}2$.}
    \label{tab:hyperparam}
  \end{minipage}
\end{table}

Cross-image aggregation is the most critical component. When the transductive adaptation pool is replaced by strictly per-image prototypes, the gain plummets from $+2.9$ to $+0.7$. This sharp drop empirically validates our core motivation (Sec.~\ref{sec:intro}): the dense prediction bottleneck of CLIP is inherently tied to the limited context within a single image, and breaking this intra-image barrier via cross-image consensus is essential for substantial improvements. Replacing DINO patch features with CLIP's own patch features reduces the gain to $+0.8$, as CLIP lacks the tight per-class spatial clustering produced by DINO's self-distillation. Crucially, removing the confidence-based anchor selection drops the gain to $+0.9$, confirming its necessity as a robust defense against confirmation bias and high-confidence hallucinations. Finally, omitting the zero-centring step reduces the gain to $+1.3$, corroborating that uncorrected anisotropic feature spaces heavily bias the late fusion.

\noindent\textbf{Sensitivity to the multiplier in $\tau = k/C$.}~~
Table~\ref{tab:hyperparam} shows that the gain is stable across $k \in [1.5, 2.5]$ (within $0.06$ of the $k{=}2$ peak at $+2.97$), confirming that TPA's performance is not sensitive to the precise value of this design constant. The gain degrades only at the extremes, where the threshold either admits near-random predictions ($k{=}1$) or excessively restricts anchors in small label-spaces ($k{=}3$).

\noindent\textbf{Choice of bank extractor.}~~
Table~\ref{tab:extractor} compares three frozen off-the-shelf encoders on NACLIP~B/16 across all eight benchmarks. All three yield positive gains, but DINOv2 outperforms MAE and SAM by a clear margin. Its self-distillation objective produces semantically consistent patch clusters that form tighter per-class prototypes than the reconstruction-based (MAE) or segmentation-guided (SAM) alternatives. The adoption of DINOv2 is therefore an empirical preference for prototype quality rather than an architectural requirement. TPA's output-level interface is independent of the auxiliary encoder, and all three alternatives still yield positive gains.

\begin{table}[h]
  \centering
  \footnotesize
  \resizebox{\linewidth}{!}{%
  \begin{tabular}{l c c c c c c c c c}
    \toprule
    \textbf{Bank extractor} & \textbf{VOC} & \textbf{Ctx} & \textbf{Obj} & \textbf{VOC20} & \textbf{Ctx59} & \textbf{Stuff} & \textbf{City} & \textbf{ADE} & \textbf{Avg.} \\
    \midrule
    SAM ViT-B/16
      & +1.0 & +1.3 & +0.8 & +1.4 & +1.5 & +1.1 & +1.9 & +1.2 & +1.3 \\
    MAE ViT-B/16
      & +1.4 & +1.8 & +1.2 & +1.9 & +2.0 & +1.5 & +2.6 & +1.7 & +1.8 \\
    \rowcolor{gray!10}
    \textbf{DINOv2 ViT-B/14 (ours)}
      & \textbf{+2.4} & \textbf{+3.1} & \textbf{+2.0} & \textbf{+3.3} & \textbf{+3.5} & \textbf{+2.6} & \textbf{+4.5} & \textbf{+2.9} & \textbf{+3.1} \\
    \bottomrule
  \end{tabular}}
  \caption{\textbf{Bank extractor ablation on NACLIP~B/16} across all benchmarks.}
  \label{tab:extractor}
  \vspace{-4mm}
\end{table}

\subsection{Composition with per-image plug-ins}
\label{sec:exp-composition}

Because TPA reads only the host's output tensor $\mathbf{p}$ and leaves the forward pass untouched, it can be stacked on top of any method that modifies the host's internals. We test this on two host$\times$plug-in pairings: (i)~ProxyCLIP~B/16 composed with FSA~\cite{fsa2025} (gradient-free attention modulation), and (ii)~NACLIP~L/14 composed with MLMP~\cite{mlmp2025} (gradient-based multi-level entropy minimisation). In each case the composed method first runs the per-image plug-in to produce an adapted output $\mathbf{p}'$, which TPA then reads as its host interface.

\begin{table}[t]
  \centering
  \footnotesize
  \caption{\textbf{Composition with per-image plug-ins.} ProxyCLIP~B/16 paired with FSA~\cite{fsa2025}; NACLIP~L/14 paired with MLMP~\cite{mlmp2025}. MLMP evaluated under its original 224$\times$224 protocol. Composed rows (\textsc{+TPA+...}) are ours.}
  \label{tab:composition}
  \resizebox{\linewidth}{!}{%
  \begin{tabular}{l l c c c c c c c c c}
    \toprule
    \textbf{Host} & \textbf{Method} & VOC & Ctx & Obj & VOC20 & Ctx59 & Stuff & City & ADE & \textbf{Avg.} \\
    \midrule
    \multirow{4}{*}{\makecell[l]{ProxyCLIP\\B/16}}
      & Base
        & 61.3 & 35.3 & 37.5 & 80.3 & 39.1 & 26.5 & 38.1 & 20.2 & 42.3 \\
    & +FSA~\cite{fsa2025}
        & 63.7 & 36.1 & 38.0 & 82.3 & 39.9 & 27.0 & 38.8 & 20.5 & 43.3\,\textcolor{BrickRed}{\scriptsize(+1.0)} \\
    & \cellcolor{gray!10}+TPA (ours)
        & \cellcolor{gray!10}62.6 & \cellcolor{gray!10}37.8 & \cellcolor{gray!10}38.7 & \cellcolor{gray!10}81.7 & \cellcolor{gray!10}41.5 & \cellcolor{gray!10}28.7 & \cellcolor{gray!10}41.2 & \cellcolor{gray!10}22.4 & \cellcolor{gray!10}\textbf{44.3}\,\textcolor{ForestGreen}{\scriptsize(+2.0)} \\
    & \cellcolor{gray!10}+TPA+FSA
        & \cellcolor{gray!10}65.4 & \cellcolor{gray!10}38.2 & \cellcolor{gray!10}41.1 & \cellcolor{gray!10}84.5 & \cellcolor{gray!10}42.0 & \cellcolor{gray!10}28.6 & \cellcolor{gray!10}41.4 & \cellcolor{gray!10}21.7 & \cellcolor{gray!10}\textbf{45.4}\,\textcolor{ForestGreen}{\scriptsize(+3.1)} \\
    \midrule
    \multirow{4}{*}{\makecell[l]{NACLIP\\L/14}}
      & Base
        & 45.1 & 25.0 & 23.8 & 75.9 & 28.2 & 18.3 & 29.5 & {---} & 35.1 \\
    & +MLMP~\cite{mlmp2025}
        & 50.8 & 28.0 & 28.8 & 83.8 & 32.0 & 21.3 & 33.4 & {---} & 39.7\,\textcolor{BrickRed}{\scriptsize(+4.6)} \\
    & \cellcolor{gray!10}+TPA (ours)
        & \cellcolor{gray!10}48.0 & \cellcolor{gray!10}30.2 & \cellcolor{gray!10}25.7 & \cellcolor{gray!10}80.0 & \cellcolor{gray!10}33.6 & \cellcolor{gray!10}22.7 & \cellcolor{gray!10}35.0 & \cellcolor{gray!10}{---} & \cellcolor{gray!10}\textbf{39.3}\,\textcolor{ForestGreen}{\scriptsize(+4.2)} \\
    & \cellcolor{gray!10}+TPA+MLMP
        & \cellcolor{gray!10}52.9 & \cellcolor{gray!10}31.9 & \cellcolor{gray!10}29.8 & \cellcolor{gray!10}86.5 & \cellcolor{gray!10}36.0 & \cellcolor{gray!10}24.4 & \cellcolor{gray!10}36.9 & \cellcolor{gray!10}{---} & \cellcolor{gray!10}\textbf{42.6}\,\textcolor{ForestGreen}{\scriptsize(+7.5)} \\
    \bottomrule
  \end{tabular}}
\end{table}

On ProxyCLIP~B/16, TPA alone ($+2.0$) already exceeds FSA ($+1.0$), and composing the two yields a further gain of $+3.1$, consistent with FSA refining spatial arrangement within each image while TPA addresses class-level errors using cross-image prototypes inaccessible to per-image methods. On NACLIP~L/14, evaluated under the MLMP evaluation protocol~\cite{mlmp2025} (224$\times$224 input, class-extended text queries), MLMP achieves $+4.6$~mIoU by updating model parameters at inference, while TPA achieves $+4.2$~mIoU without updating any parameters. Operating solely on the host's output tensor $\mathbf{p}$ and leaving the forward pass untouched, TPA is decoupled from the host's internals and remains fully composable with adaptation methods that modify internal representations. Stacking TPA on top of MLMP yields $+7.5$~mIoU, exceeding each method individually. The combined gain falls below the na\"ive sum of the two deltas, reflecting a partial overlap in the corrected pixels. However, the substantial net improvement confirms that the two methods target orthogonal error modes: MLMP refines intra-image spatial coherence via internal weight updates, while TPA corrects global class-level confusion via inter-image priors. This seamless composability firmly establishes TPA's nature as an architecture-agnostic plug-in, fully compatible with existing test-time adaptation pipelines.

\subsection{Bank size and efficiency}
\label{sec:exp-efficiency}

\begin{figure*}[t]
  \centering
  \includegraphics[width=\textwidth]{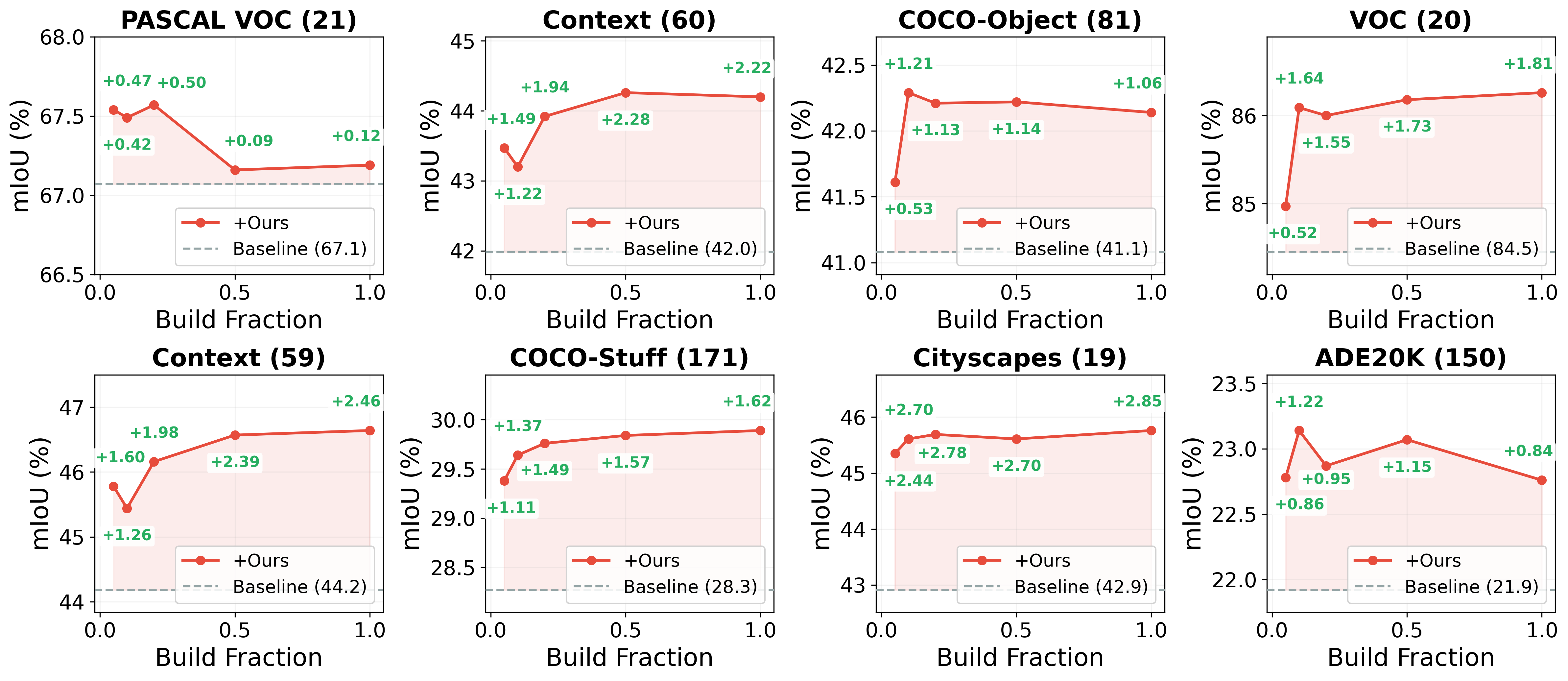}
  \caption{Bank-size saturation on Trident~B/16 across subset fractions.}
  \label{fig:saturation-main}
\end{figure*}

Figure~\ref{fig:saturation-main} shows that on most benchmarks the gain saturates within approximately 10\% of the adaptation pool, with the largest marginal returns coming from the first few percent of images where the rarest classes first accumulate prototypes. Semantically dense datasets (e.g.\ Context59) continue to improve modestly beyond this point, reflecting greater within-class visual diversity at scale. Either way, TPA does not require a large or carefully curated adaptation set.

\begin{figure*}[t]
  \centering
  \includegraphics[width=\textwidth]{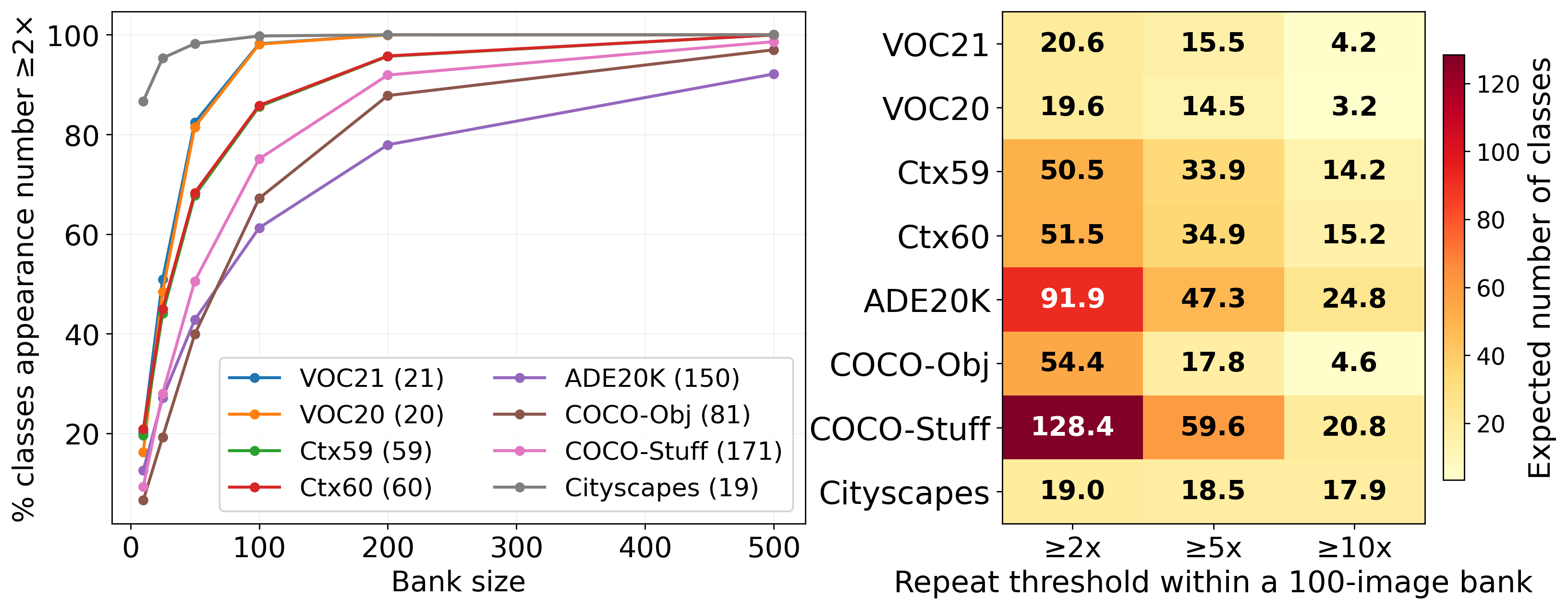}
  \caption{\textbf{Class repetition vs bank size.} Left: \% of classes appearing $\ge2\times$. Right: expected class counts at various repeat thresholds within a 100-image bank.}
  \label{fig:repeat}
\end{figure*}

A natural concern is whether 100 images are enough for every category to accumulate multiple representative anchors. Figure~\ref{fig:repeat} addresses this directly. A 100-image bank corresponds to 1--10\% of the adaptation pool for most benchmarks (up to 20\% for Cityscapes with its compact 500-image val set). Within this bank, virtually every class has appeared at least twice and the majority five or more times, letting TPA average out per-image noise in the DINOv2 feature space and produce stable class centroids. The rapid class repetition explains why the mIoU gain saturates early, well before the full pool is consumed.

\begin{figure*}[t]
  \centering
  \includegraphics[width=\textwidth]{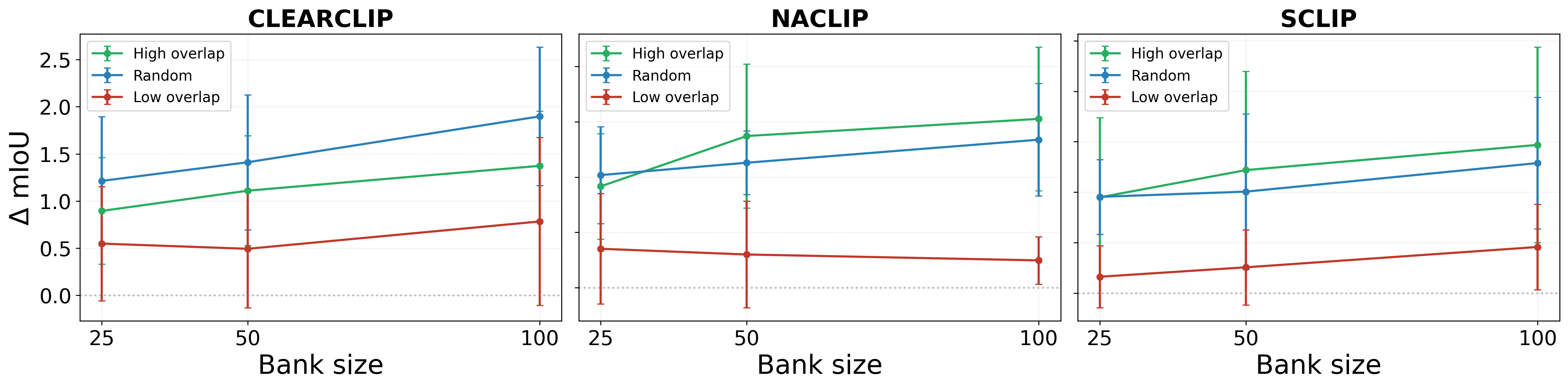}
  \caption{\textbf{Label-overlap robustness.} Per-class $\Delta$\,mIoU under forced high, random, and low overlap between bank and evaluation labels. Gains remain non-negative even under adversarial low-overlap splits.}
  \label{fig:overlap}
\end{figure*}

A related question is whether TPA requires the bank and evaluation set to share the same semantic classes. We construct bank/evaluation splits with forced high, random, or low overlap in label space (Fig.~\ref{fig:overlap}). Under the realistic random-overlap condition, gains remain consistently positive. More importantly, under the adversarial low-overlap split---where the bank misses most of the evaluated classes---the gain shrinks but rigorously stays non-negative. This empirically validates the safety of our late-fusion design (Eq.~\ref{eq:fusion}): for classes without valid prototypes, the auxiliary signal naturally vanishes, allowing the model to fall back to the host's original calibration without penalizing unseen categories (bounded degradation). The largest gain predictably occurs under high overlap, as frequently appearing classes accumulate more precise centroids.

\noindent\textbf{Computational cost.}~~
Table~\ref{tab:efficiency} details TPA's overhead relative to the ProxyCLIP host. The bank is built \emph{once} offline; each image's DINOv2 patch features are extracted and cached in CPU memory (fp16). At inference, ProxyCLIP's existing DINOv2 pass is re-used; TPA adds only a CPU$\to$GPU transfer and cosine lookup (0.33--0.45\,ms). Hosts without a built-in DINOv2 encoder additionally incur one frozen DINOv2 forward pass per test image, adding overhead comparable to the DINOv2 column in Table~\ref{tab:efficiency}. The prototype bank totals at most 225\,KB (ADE20K, fp16).

\begin{table*}[t]
  \caption{\textbf{Computational overhead of TPA.} Benchmarked on an A100 GPU. TPA offline bank build is amortized; inference adds only a lightweight cosine lookup.}
  \label{tab:efficiency}
  \centering
  \small
  \resizebox{\textwidth}{!}{%
  \setlength{\tabcolsep}{5pt}
  \begin{tabular}{l c c c c c c c c c}
    \toprule
    & \multicolumn{3}{c}{\textbf{ProxyCLIP base (ms)}} & \multicolumn{2}{c}{\textbf{Bank build (amortised)}} & \multicolumn{2}{c}{\textbf{Inference overhead (ms)}} & \multicolumn{2}{c}{\textbf{Bank size (fp16)}} \\
    \cmidrule(lr){2-4}\cmidrule(lr){5-6}\cmidrule(lr){7-8}\cmidrule(lr){9-10}
    \textbf{Backbone} & CLIP & DINOv2 & \textbf{Total} & DINO pass & Update & Feat.\ transfer & Lookup & Prototype & Cache/img \\
    \midrule
    ViT-B/16 & 6.6 & 5.7 & 12.3 & 5.8\,ms & $<$0.5\,ms & \multirow{3}{*}{$<$0.1\,ms} & 0.33\,ms & \multirow{3}{*}{\makecell{32--225\,KB\\(VOC--ADE)}} & 0.4\,MB \\
    ViT-L/14 & 13.0 & 8.0 & 21.0 & 8.0\,ms & $<$0.5\,ms & & 0.40\,ms & & 0.9\,MB \\
    ViT-H/14 & 22.1 & 11.2 & 33.2 & 11.2\,ms & $<$0.5\,ms & & 0.45\,ms & & 1.5\,MB \\
    \bottomrule
  \end{tabular}}
\end{table*}

\subsection{Qualitative results}
\label{sec:exp-qualitative}

\begin{figure*}[t]
  \centering
  \includegraphics[width=0.9\textwidth]{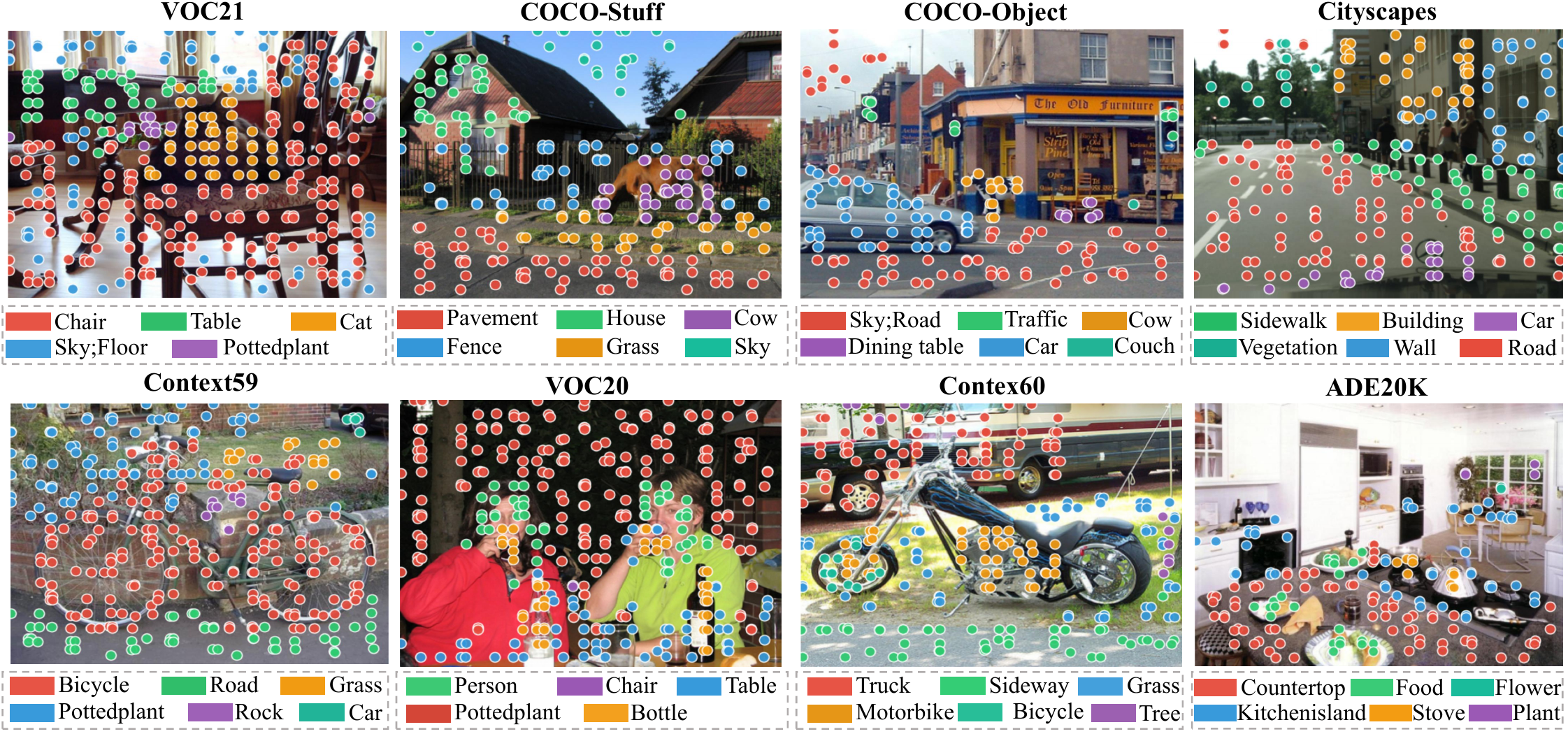}
  \caption{\textbf{Per-class anchor regions (ProxyCLIP host).} Coloured dots visualize pixels satisfying the anchor criterion (Eq.~\eqref{eq:anchor-rule}). Label-space sizes range from 19 to 164.}
  \label{fig:anchor-vis}
\end{figure*}

Figure~\ref{fig:anchor-vis} shows, for one image per benchmark, the spatial distribution of pixels satisfying the anchor criterion (Eq.~\eqref{eq:anchor-rule}). Across all eight benchmarks, despite the zero-shot nature of the host, the strictly filtered high-confidence pixels consistently form spatially coherent, class-specific clusters that perfectly match the semantic layout of each scene. This visualization directly corroborates the effectiveness of our joint defense mechanism introduced in Sec.~\ref{sec:method-pass1}: it successfully circumvents confirmation bias by rejecting noisy or hallucinated predictions. This precise localization guarantees the high quality of the subsequent prototype construction (Eq.~\eqref{eq:stats}), as averaging DINO features over these reliable regions produces tight, highly discriminative prototypes.

\begin{figure}[H]
  \centering
  \includegraphics[width=0.9\textwidth]{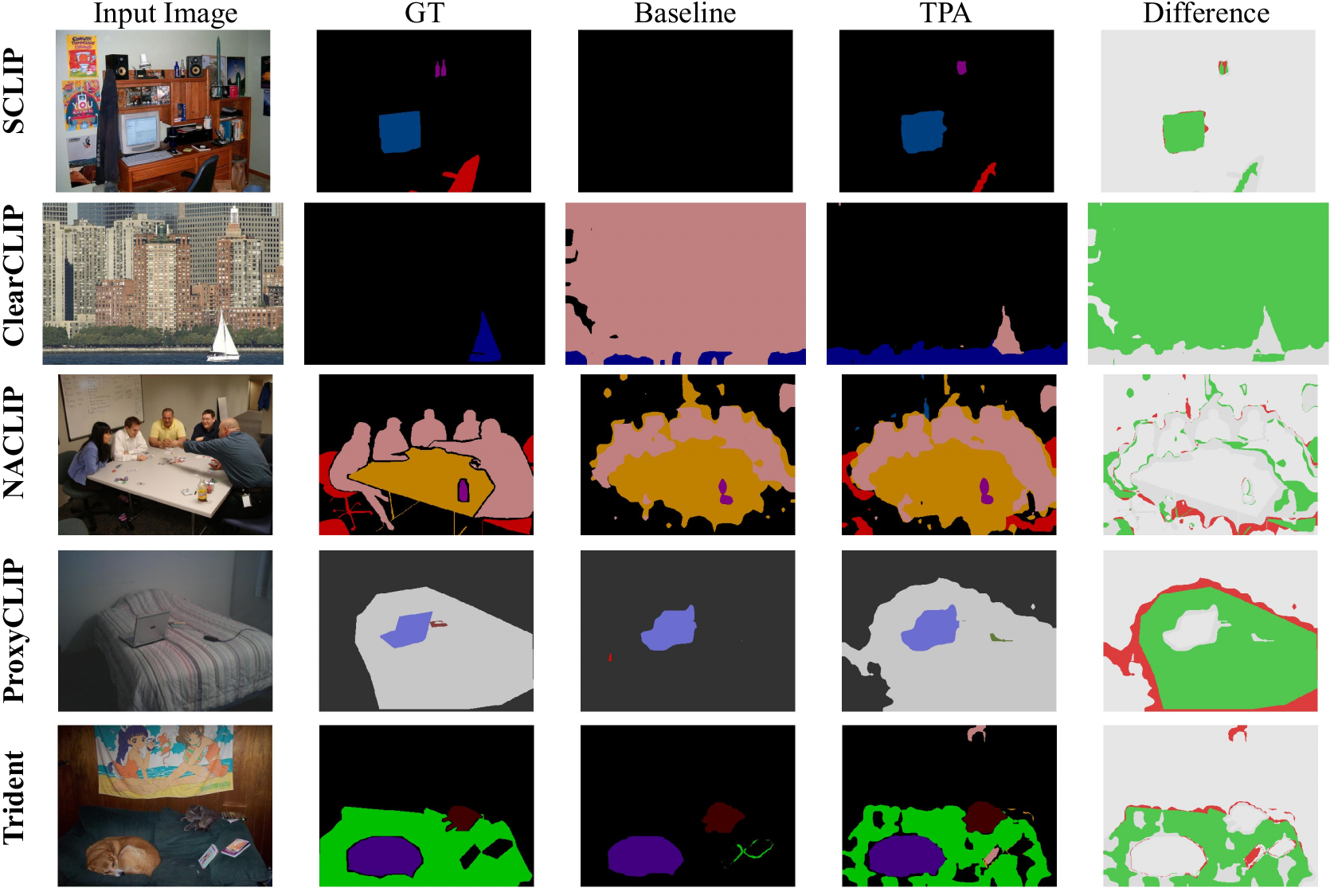}
  \caption{\textbf{Qualitative comparison across hosts on VOC21.} Images with the largest per-image mIoU gain. Difference maps highlight recovered regions in green, regressions in red, and pixels correctly predicted by both methods in light grey.}
  \label{fig:qualitative}
\end{figure}

Figure~\ref{fig:qualitative} shows that across all five hosts, baseline predictions often miss a class or fragment its regions, whereas TPA recovers coherent semantic regions using prototypes aggregated over the adaptation pool. The difference maps localize improvements to semantic regions rather than scattered pixels, consistent with the spatial coherence of DINO features.

%% file: sections/arxiv_conclusion.tex
\section{Conclusion}
\label{sec:conclusion}

We present TPA, a training-free plug-in for open-vocabulary semantic segmentation that aggregates cross-image class prototypes from an unlabeled deployment pool and fuses them with any host's logits via a cosine lookup, without labels, gradient updates, or host modifications.  Across ten host$\times$backbone configurations and eight benchmarks, TPA yields consistent positive gains, complementing per-image spatial methods and composing additively with parameter-adapting approaches such as MLMP ($+7.5$\,mIoU combined on NACLIP~L/14).  Gains concentrate on difficult classes, remain non-negative even under adversarially low label overlap, and saturate with small prototype banks (about 10\% of images), keeping additional overhead modest when reusing hosts with DINOv2 and otherwise requiring a single frozen DINOv2 pass per image.  Operating purely at the output level makes TPA portable across hosts and straightforward to stack with internal adaptation methods.  We hope these results motivate further exploration of the test stream as a source of implicit supervision for open-vocabulary dense prediction.

%% file: sections/arxiv_appendix.tex
\section{Algorithm}
\label{sec:appendix-algo}

Algorithms~\ref{alg:adapt} and~\ref{alg:infer} give complete pseudocode for the two phases of TPA.
All symbols follow Section~3 of the main paper.

\begin{algorithm}[H]
\caption{TPA --- Adaptation Phase}
\label{alg:adapt}
\begin{algorithmic}[1]
\Require Pool $\mathcal{P}=\{I_1,\dots,I_M\}$; frozen host $f$; frozen DINO encoder $g$;
         class set $\mathcal{C}=\{1,\dots,C\}$; threshold $\tau{=}2/C$; min-anchor count $K_{\min}$
\Ensure  Unit-norm prototype bank $\{\mathbf{P}_c\}_{c=1}^{C}$, covered set $\mathcal{C}^{\star}$
\State $\mathbf{s}_c \leftarrow \mathbf{0},\; N_c \leftarrow 0$ \quad for all $c \in \mathcal{C}$
\For{$n = 1,\dots,M$}
  \State $\mathbf{p} \leftarrow f(I_n;\mathcal{C})$ \Comment{dense class probabilities, shape $C{\times}H{\times}W$}
  \State $\boldsymbol{\phi} \leftarrow g(I_n)$ \Comment{DINO patch features, shape $D{\times}H{\times}W$}
  \For{$c \in \mathcal{C}$}
    \State $\mathcal{A}^c \leftarrow \bigl\{(i,j) : \arg\max_{c'}\mathbf{p}(c',i,j)=c \;\wedge\; \mathbf{p}(c,i,j)>\tau\bigr\}$
    \If{$|\mathcal{A}^c| \geq K_{\min}$}
      \State $\mathbf{s}_c \mathrel{+}= \sum_{(i,j)\in\mathcal{A}^c}\boldsymbol{\phi}(i,j)$;\quad $N_c \mathrel{+}= |\mathcal{A}^c|$
    \EndIf
  \EndFor
\EndFor
\State $\mathbf{P}_c \leftarrow \mathbf{s}_c/\lVert\mathbf{s}_c\rVert_2$ if $N_c>0$, else $\mathbf{0}$;\quad
       $\mathcal{C}^{\star} \leftarrow \{c:N_c>0\}$
\State \textbf{return} $\{\mathbf{P}_c\}$, $\mathcal{C}^{\star}$
\end{algorithmic}
\end{algorithm}

\begin{algorithm}[H]
\caption{TPA --- Inference}
\label{alg:infer}
\begin{algorithmic}[1]
\Require Test image $I^*$; host $f$; encoder $g$; bank $\{\mathbf{P}_c\}$, $\mathcal{C}^{\star}$;
         fusion weight $\alpha$; temperature $\lambda$
\Ensure  Dense label map $\hat{y}$
\State $\mathbf{p}^* \leftarrow f(I^*;\mathcal{C})$;\quad $\boldsymbol{\phi}^* \leftarrow g(I^*)$
\For{each pixel $(i,j)$}
  \State $\hat{\boldsymbol{\phi}} \leftarrow \boldsymbol{\phi}^*(i,j)/\lVert\boldsymbol{\phi}^*(i,j)\rVert_2$
  \State $\mathbf{q}(c,i,j) \leftarrow \mathbf{P}_c^{\top}\hat{\boldsymbol{\phi}}\cdot\mathbb{1}[c\in\mathcal{C}^{\star}]$ \quad $\forall c$ \Comment{cosine scores}
  \State $\bar{q}(i,j) \leftarrow \tfrac{1}{|\mathcal{C}^{\star}|}\sum_{c\in\mathcal{C}^{\star}}\mathbf{q}(c,i,j)$ \Comment{per-pixel mean over prototypes}
  \State $\tilde{\mathbf{q}}(c,i,j) \leftarrow \bigl(\mathbf{q}(c,i,j)-\bar{q}(i,j)\bigr)\cdot\mathbb{1}[c\in\mathcal{C}^{\star}]$ \Comment{mean-centred}
  \State $\ell(c,i,j) \leftarrow \log\mathbf{p}^*(c,i,j) + (1-\alpha)\lambda\,\tilde{\mathbf{q}}(c,i,j)$
\EndFor
\State $\hat{y}(i,j) \leftarrow \arg\max_c\;\ell(c,i,j)$ \quad for all $(i,j)$
\State \textbf{return} $\hat{y}$
\end{algorithmic}
\end{algorithm}

Three properties follow directly. \emph{Host-agnostic}: only $\mathbf{p}$ is read from $f$ (line~3); no internal attention, hidden state, or gradient is accessed. \emph{Label- and gradient-free}: no ground-truth label enters either algorithm. \emph{Incremental}: the running sum in line~8 admits streaming updates without revisiting previous images, so the bank can be extended online.

\section{Extended Results and Backbone Scaling Analysis}
\label{sec:appendix-tables}

In the main paper, due to space constraints, our primary evaluation focused on the ViT-B/16 backbone. This section provides the exhaustive quantitative evaluation on the larger ViT-L/14 and ViT-H/14 backbones. Furthermore, we provide a detailed analysis of how TPA's performance scales with the host model's capacity and internal Vision Foundation Model (VFM) choices.

\subsection{Performance on Larger Capacity Backbones}

Table~\ref{tab:main-full} reports full results for all L/14 and H/14 backbone configurations. We observe a consistent trend: scaling up the backbone capacity from ViT-B/16 to ViT-L/14 and ViT-H/14 uniformly increases the baseline performance of all host architectures. Remarkably, despite the stronger baselines, TPA continues to deliver substantial and orthogonal performance gains across the board. 

For instance, on the SCLIP architecture, TPA achieves an impressive $+8.3$ mIoU gain on the L/14 backbone, significantly outperforming the internal modification baseline (+6.0). This verifies that the information bottleneck inherent in isolated single-image inference persists even as the parameter count of the CLIP backbone increases. TPA's cross-image consensus mechanism successfully addresses this bottleneck regardless of the underlying model scale. Table~\ref{tab:resclip-naclip} provides the specific ResCLIP baseline on NACLIP~B/16 for completeness.

\begin{table*}[t]
  \caption{\textbf{Supplementary main results} (L/14 and H/14 backbones; B/16 in Table~1 of the main paper).  Same notation;
  \textsc{+FSA} rows from~\cite{fsa2025}; \textsc{+MLMP} row from~\cite{mlmp2025}.}
  \label{tab:main-full}
  \centering
  \footnotesize
  \resizebox{\linewidth}{!}{%
  \begin{tabular}{l l c c c c c c c c c}
    \toprule
    \textbf{Host / Method} & \textbf{Backbone} &
    VOC & Ctx & Obj & VOC20 & Ctx59 & Stuff & City & ADE & \textbf{Avg.} \\
    \midrule
    SCLIP~\cite{sclip2024}  & L/14
      & 43.5 & 22.3 & 25.0 & 69.1 & 25.2 & 17.6 & 18.6 & 10.9 & 29.0 \\
    \quad +FSA~\cite{fsa2025} & L/14
      & 48.1 & 27.8 & 30.8 & 79.9 & 30.3 & 20.4 & 27.1 & 15.9 & 35.0\,\textcolor{BrickRed}{\scriptsize(+6.0)} \\
    \rowcolor{gray!10}
    \quad +Ours & L/14
      & 51.3 & 29.8 & 32.5 & 84.1 & 33.7 & 23.0 & 30.3 & 13.7 & \textbf{37.3}\,\textcolor{ForestGreen}{\scriptsize(+8.3)} \\
    \midrule
    ClearCLIP~\cite{clearclip2024} & L/14
      & 46.1 & 29.6 & 26.7 & 80.0 & 30.1 & 19.9 & 27.9 & 15.0 & 34.4 \\
    \quad +FSA~\cite{fsa2025} & L/14
      & 47.5 & 30.8 & 27.9 & 80.4 & 30.2 & 20.4 & 27.2 & 16.8 & 35.2\,\textcolor{BrickRed}{\scriptsize(+0.8)} \\
    \rowcolor{gray!10}
    \quad +Ours & L/14
      & 47.6 & 31.3 & 32.0 & 83.2 & 33.1 & 22.8 & 33.5 & 17.6 & \textbf{37.6}\,\textcolor{ForestGreen}{\scriptsize(+3.2)} \\
    \midrule
    NACLIP~\cite{naclip2024} & L/14
      & 57.9 & 33.7 & 34.4 & 84.5 & 36.1 & 23.7 & 34.5 & 18.8 & 40.4 \\
    \quad +MLMP~\cite{mlmp2025} & L/14
      & 63.6 & 36.7 & 39.4 & {---} & 39.8 & 26.6 & 38.4 & {---} & {---} \\
    \rowcolor{gray!10}
    \quad +Ours & L/14
      & 59.9 & 38.1 & 35.5 & 87.6 & 40.7 & 27.3 & 39.1 & 22.4 & \textbf{43.8}\,\textcolor{ForestGreen}{\scriptsize(+3.3)} \\
    \midrule
    ProxyCLIP~\cite{proxyclip2024} & L/14
      & 60.6 & 34.5 & 39.2 & 83.2 & 37.7 & 25.6 & 40.1 & 22.6 & 42.9 \\
    \quad +FSA~\cite{fsa2025} & L/14
      & 61.8 & 34.9 & 40.2 & 84.1 & 38.1 & 25.9 & 41.2 & 22.9 & \textbf{43.6}\,\textcolor{BrickRed}{\scriptsize(+0.7)} \\
    \rowcolor{gray!10}
    \quad +Ours & L/14
      & 61.1 & 36.6 & 38.0 & 84.4 & 40.5 & 27.9 & 42.6 & 24.6 & \textbf{44.5}\,\textcolor{ForestGreen}{\scriptsize(+1.6)} \\
    ProxyCLIP~\cite{proxyclip2024} & H/14
      & 65.0 & 35.4 & 38.6 & 83.3 & 39.6 & 26.8 & 42.0 & 24.2 & 44.4 \\
    \quad +FSA~\cite{fsa2025} & H/14
      & 67.9 & 36.3 & 40.2 & 85.7 & 40.5 & 27.3 & 43.6 & 24.5 & \textbf{45.8}\,\textcolor{ForestGreen}{\scriptsize(+1.4)} \\
    \rowcolor{gray!10}
    \quad +Ours & H/14
      & 64.4 & 37.1 & 38.1 & 83.7 & 41.9 & 28.3 & 43.7 & 25.6 & 45.4\,\textcolor{BrickRed}{\scriptsize(+1.0)} \\
    \midrule
    Trident~\cite{trident2024} & H/14
      & 70.5 & 44.0 & 41.7 & 88.2 & 46.7 & 28.7 & 46.2 & 26.3 & 49.0 \\
    \rowcolor{gray!10}
    \quad +Ours & H/14
      & 69.8 & 45.1 & 41.8 & 90.0 & 48.4 & 30.8 & 48.5 & 27.7 & \textbf{50.3}\,\textcolor{ForestGreen}{\scriptsize(+1.3)} \\
    \bottomrule
  \end{tabular}}
\end{table*}

\begin{table}[t]
  \caption{\textbf{ResCLIP~\cite{resclip2025} on NACLIP~B/16.}  \textsc{+ResCLIP} from our own evaluation;
  \cite{resclip2025} does not report NACLIP.}
  \label{tab:resclip-naclip}
  \centering
  \footnotesize
  \resizebox{\linewidth}{!}{%
  \begin{tabular}{l c c c c c c c c c}
    \toprule
    \textbf{Method} & VOC & Ctx & Obj & VOC20 & Ctx59 & Stuff & City & ADE & \textbf{Avg.} \\
    \midrule
    NACLIP B/16 & 64.1 & 37.7 & 36.2 & 83.0 & 40.0 & 25.7 & 38.3 & 19.1 & 43.0 \\
    \quad+ResCLIP~\cite{resclip2025} & 65.0 & 38.2 & 37.4 & 85.0 & 41.0 & 26.5 & 39.0 & 19.8 & 44.0\,\textcolor{BrickRed}{\scriptsize(+1.0)} \\
    \rowcolor{gray!10}
    \quad+Ours & 66.5 & 40.8 & 38.2 & 86.3 & 43.5 & 28.3 & 42.8 & 22.0 & \textbf{46.0}\,\textcolor{ForestGreen}{\scriptsize(+3.0)} \\
    \bottomrule
  \end{tabular}}
\end{table}

\subsection{Robustness to Internal VFM Selection}

Table~\ref{tab:vfm-full} extends the VFM generalisation experiment to the L/14 and H/14 backbones. Similar to our findings on the B/16 architecture, the results strictly maintain the property that TPA's external DINOv2 prototype bank complements a wide array of internally injected VFMs (SAM, MAE, DINOv2). The improvement margins are particularly pronounced when the host relies on MAE features (+2.6 on L/14, +3.0 on H/14). This disparity reinforces the hypothesis that while MAE excels at reconstructive representation learning, its localized patch descriptors lack the explicit semantic alignment that DINOv2 builds through self-distillation. TPA flawlessly bridges this representation gap at test time via its decoupled late-fusion pipeline.

\begin{table*}[t]
  \caption{\textbf{VFM generalisation: L/14 and H/14 backbones} (B/16 in Table~2 of the main paper).
  TPA bank extractor (DINOv2 ViT-B/14) held fixed; only ProxyCLIP's internal proxy varies.
  Proxy and \textsc{+FSA} rows from~\cite{fsa2025}.}
  \label{tab:vfm-full}
  \centering
  \footnotesize
  \resizebox{\linewidth}{!}{%
  \begin{tabular}{l l l c c c c c c c c c}
    \toprule
    \textbf{CLIP} & \textbf{VFM} & \textbf{Method} &
    VOC & Ctx & Obj & VOC20 & Ctx59 & Stuff & City & ADE & \textbf{Avg.} \\
    \midrule
    \multirow{9}{*}{\rotatebox{90}{ViT-L/14}}
    & \multirow{3}{*}{\makecell{SAM\\ViT-B/16}}
      & Proxy                    & 57.2 & 32.6 & 36.5 & 82.3 & 35.6 & 24.2 & 39.1 & 20.7 & 41.0 \\
    && \quad+FSA~\cite{fsa2025}  & 58.5 & 33.0 & 38.0 & 83.1 & 36.1 & 24.5 & 40.5 & 21.0 & 41.8\,\textcolor{BrickRed}{\scriptsize(+0.8)} \\
    && \cellcolor{gray!10}\quad+Ours & \cellcolor{gray!10}57.5 & \cellcolor{gray!10}32.8 & \cellcolor{gray!10}37.1 & \cellcolor{gray!10}85.0 & \cellcolor{gray!10}35.9 & \cellcolor{gray!10}26.8 & \cellcolor{gray!10}42.8 & \cellcolor{gray!10}22.5 & \cellcolor{gray!10}\textbf{42.6}\,\textcolor{ForestGreen}{\scriptsize(+1.6)} \\
    \cmidrule(lr){2-12}
    & \multirow{3}{*}{\makecell{MAE\\ViT-B/16}}
      & Proxy                    & 49.0 & 27.8 & 31.6 & 78.3 & 30.2 & 20.8 & 31.8 & 17.2 & 35.8 \\
    && \quad+FSA~\cite{fsa2025}  & 52.8 & 29.9 & 34.9 & 80.2 & 32.5 & 22.6 & 34.7 & 19.0 & 38.3\,\textcolor{BrickRed}{\scriptsize(+2.5)} \\
    && \cellcolor{gray!10}\quad+Ours & \cellcolor{gray!10}50.4 & \cellcolor{gray!10}28.7 & \cellcolor{gray!10}33.5 & \cellcolor{gray!10}82.5 & \cellcolor{gray!10}31.5 & \cellcolor{gray!10}23.7 & \cellcolor{gray!10}37.0 & \cellcolor{gray!10}19.5 & \cellcolor{gray!10}\textbf{38.4}\,\textcolor{ForestGreen}{\scriptsize(+2.6)} \\
    \cmidrule(lr){2-12}
    & \multirow{3}{*}{\makecell{DINOv2\\ViT-B/14}}
      & Proxy                    & 56.6 & 33.0 & 36.7 & 85.2 & 36.2 & 24.6 & 35.2 & 21.6 & 41.1 \\
    && \quad+FSA~\cite{fsa2025}  & 57.4 & 33.3 & 38.0 & 85.8 & 36.5 & 24.7 & 36.1 & 21.9 & 41.7\,\textcolor{BrickRed}{\scriptsize(+0.6)} \\
    && \cellcolor{gray!10}\quad+Ours & \cellcolor{gray!10}58.5 & \cellcolor{gray!10}34.1 & \cellcolor{gray!10}38.8 & \cellcolor{gray!10}86.5 & \cellcolor{gray!10}37.6 & \cellcolor{gray!10}25.8 & \cellcolor{gray!10}37.9 & \cellcolor{gray!10}22.8 & \cellcolor{gray!10}\textbf{42.8}\,\textcolor{ForestGreen}{\scriptsize(+1.7)} \\
    \midrule
    \multirow{9}{*}{\rotatebox{90}{ViT-H/14}}
    & \multirow{3}{*}{\makecell{SAM\\ViT-B/16}}
      & Proxy                    & 63.5 & 34.1 & 36.7 & 84.0 & 37.9 & 25.0 & 41.1 & 22.0 & 43.1 \\
    && \quad+FSA~\cite{fsa2025}  & 64.9 & 34.4 & 37.6 & 85.5 & 38.0 & 25.1 & 42.6 & 21.9 & 43.7\,\textcolor{BrickRed}{\scriptsize(+0.6)} \\
    && \cellcolor{gray!10}\quad+Ours & \cellcolor{gray!10}63.8 & \cellcolor{gray!10}34.2 & \cellcolor{gray!10}38.2 & \cellcolor{gray!10}86.5 & \cellcolor{gray!10}38.1 & \cellcolor{gray!10}27.7 & \cellcolor{gray!10}45.8 & \cellcolor{gray!10}24.0 & \cellcolor{gray!10}\textbf{44.8}\,\textcolor{ForestGreen}{\scriptsize(+1.7)} \\
    \cmidrule(lr){2-12}
    & \multirow{3}{*}{\makecell{MAE\\ViT-B/16}}
      & Proxy                    & 54.7 & 29.8 & 32.2 & 80.6 & 32.9 & 21.8 & 34.9 & 19.4 & 38.3 \\
    && \quad+FSA~\cite{fsa2025}  & 58.7 & 32.0 & 35.2 & 82.2 & 35.3 & 24.0 & 37.9 & 21.3 & 40.8\,\textcolor{BrickRed}{\scriptsize(+2.5)} \\
    && \cellcolor{gray!10}\quad+Ours & \cellcolor{gray!10}57.1 & \cellcolor{gray!10}31.9 & \cellcolor{gray!10}34.5 & \cellcolor{gray!10}84.4 & \cellcolor{gray!10}35.7 & \cellcolor{gray!10}25.0 & \cellcolor{gray!10}40.1 & \cellcolor{gray!10}21.4 & \cellcolor{gray!10}\textbf{41.3}\,\textcolor{ForestGreen}{\scriptsize(+3.0)} \\
    \cmidrule(lr){2-12}
    & \multirow{3}{*}{\makecell{DINOv2\\ViT-B/14}}
      & Proxy                    & 61.5 & 34.0 & 37.3 & 86.1 & 37.8 & 26.2 & 37.8 & 23.4 & 43.0 \\
    && \quad+FSA~\cite{fsa2025}  & 63.0 & 34.4 & 38.5 & 87.4 & 38.4 & 26.4 & 39.7 & 23.7 & 43.9\,\textcolor{BrickRed}{\scriptsize(+0.9)} \\
    && \cellcolor{gray!10}\quad+Ours & \cellcolor{gray!10}64.1 & \cellcolor{gray!10}35.2 & \cellcolor{gray!10}39.4 & \cellcolor{gray!10}88.1 & \cellcolor{gray!10}39.5 & \cellcolor{gray!10}27.5 & \cellcolor{gray!10}41.0 & \cellcolor{gray!10}24.8 & \cellcolor{gray!10}\textbf{44.9}\,\textcolor{ForestGreen}{\scriptsize(+1.9)} \\
    \bottomrule
  \end{tabular}}
\end{table*}

\section{Per-class analysis}
\label{sec:appendix-perclass}

\begin{figure*}[t]
  \centering
  \includegraphics[width=\textwidth]{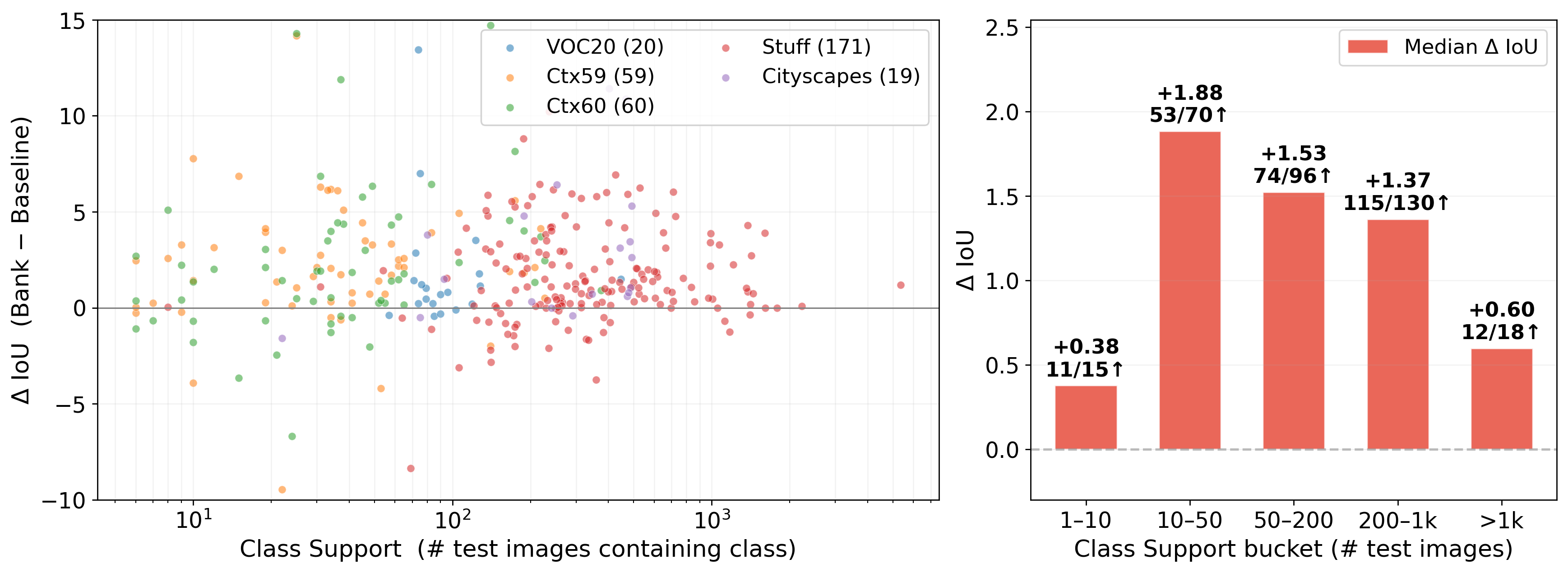}
  \caption{\textbf{Per-class $\Delta$\,mIoU vs.\ class support.}  Each point is one semantic class;
  support = number of test images in which it appears.  Three representative host$\times$dataset pairs are shown.
  Rare classes (low support) remain predominantly positive, confirming that prototype quality degrades gracefully as class frequency decreases.}
  \label{fig:appendix-support}
\end{figure*}

\begin{figure*}[t]
  \centering
  \includegraphics[width=0.92\textwidth]{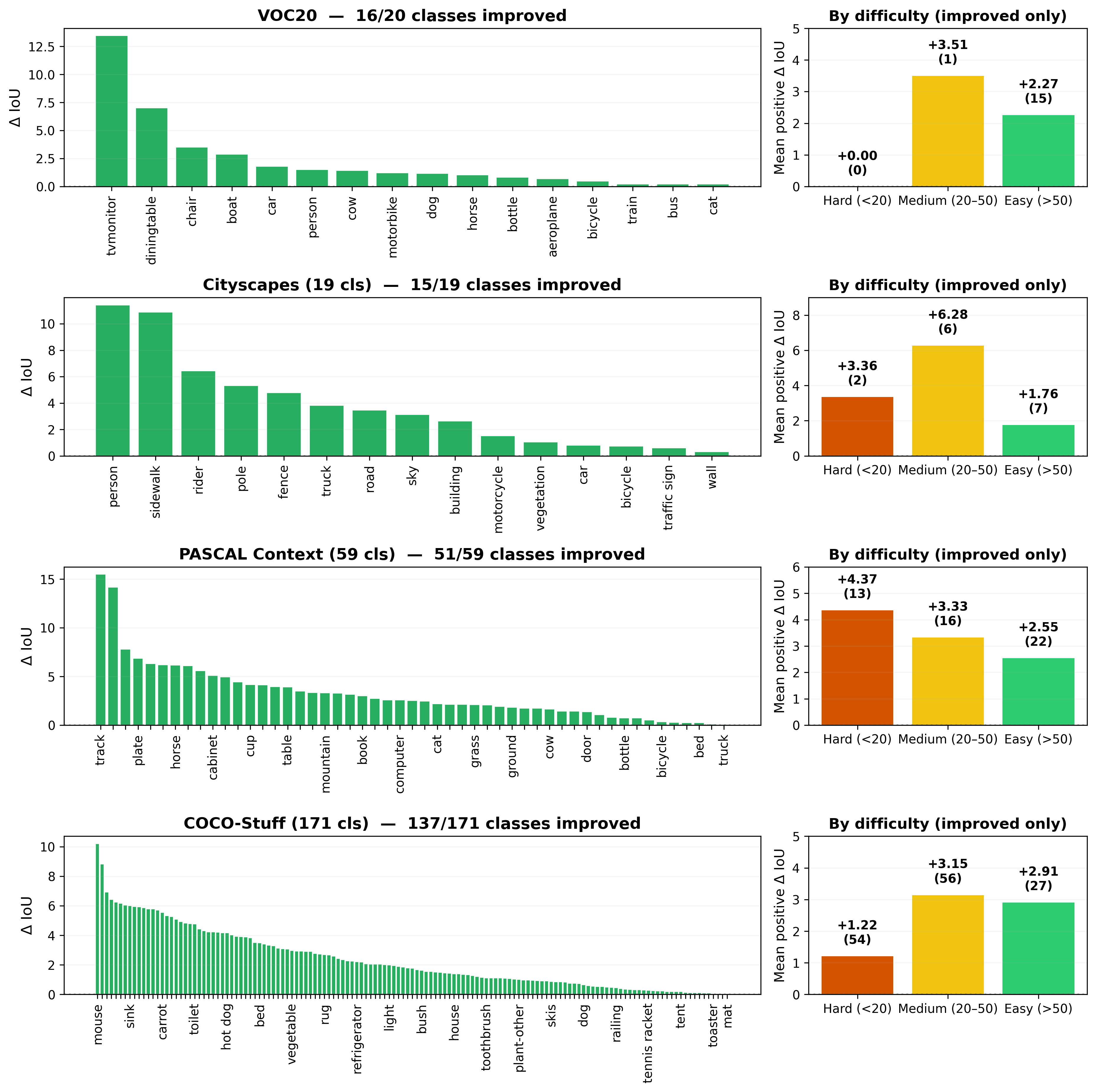}
  \caption{\textbf{Per-class $\Delta$\,IoU on Trident~B/16.}  Classes sorted by baseline IoU.
  The largest positive shifts concentrate on classes with low baseline IoU, indicating that TPA preferentially corrects the host's most uncertain predictions.}
  \label{fig:appendix-per-class}
\end{figure*}

TPA's anchor-selection criterion retains only high-confidence, argmax-consistent patches, so the bank for rare classes is built from their most canonical appearances.
Figure~\ref{fig:appendix-support} shows that this is sufficient: even classes appearing in fewer than 50 adaptation-pool images yield positive average $\Delta$\,mIoU. This empirically validates our claim in the main text: the strict dual-condition filtering ensures that even a tiny subset of images captures highly canonical, noise-free prototypes.

Figure~\ref{fig:appendix-per-class} shows a consistent pattern across individual classes on Trident~B/16: classes the host struggles with most (low baseline IoU) benefit most from the prototype-based correction. The cross-image consensus acts as a robust regularizer precisely where intra-image context fails.

\section{Discussion}
\label{sec:appendix-discussion}

This section elaborates on design choices, evaluation protocol, and anticipated reviewer questions that space constraints prevented us from addressing fully in the main paper.

\paragraph{Fairness of the Transductive Protocol.}
TPA follows the transductive test-time adaptation protocol of T3A~\cite{t3a2021} and LaMe~\cite{lame2022}: before inference, the method has access to the \emph{unlabeled} deployment images, and the goal is to improve predictions on those same images.
A natural question is whether this gives TPA an unfair advantage over purely inductive baselines such as SCLIP, NACLIP, or ClearCLIP, which see each test image only once.

We argue it does not, for three reasons.
First, no ground-truth label is accessed at any point: the anchor criterion (Sec.~3.1 of the main paper) uses only the host's own output probabilities, and the prototype bank stores only DINO features.
Second, the gain is measured on the full evaluation partition; TPA does not tune itself to specific test images but constructs class-level statistics over a separate 100-image adaptation pool drawn from the same domain.
Third, test-time use of unlabeled deployment images is a standard and explicitly recognised protocol in the TTA literature~\cite{t3a2021,lame2022,mlmp2025}; MLMP~\cite{mlmp2025}, which TPA outperforms without updating any parameters, performs substantially more invasive test-time adaptation via gradient-based entropy minimisation.
The fair comparison is therefore TPA vs.\ parameter-updating TTA methods (favourable to TPA) or vs.\ inductive training-free methods with the understanding that TPA assumes access to a small unlabeled pool (explicit assumption, stated in Sec.~2 of the main paper).

\paragraph{Correlation between Host Quality and TPA Gains.}
TPA's margin is largest on attention-redesign hosts (SCLIP, NACLIP, ClearCLIP, average $+3.3$) and smallest on VFM-injection hosts (ProxyCLIP, Trident, average $+1.7$).
This inverse relationship with host quality is expected and mechanistically transparent.
Attention-redesign hosts improve CLIP's spatial behaviour but do so purely from per-image context; their patch predictions remain noisy for classes that appear in unusual viewpoints or low contrast.
VFM-injection hosts additionally inject per-image DINOv2 or SAM priors, which already provide strong local spatial priors.
TPA's cross-image aggregation provides a complementary signal: class-level visual statistics accumulated over many images that are inaccessible to any per-image method.
For a host that already localises regions well, TPA's marginal correction is necessarily smaller.
Critically, gains remain positive for every host and every backbone (Table~\ref{tab:main-full}), confirming that cross-image prototypes carry signal over and above per-image priors regardless of how good those priors are.

\paragraph{The ProxyCLIP H/14 Exception.}
The one configuration where TPA ($+1.0$) trails FSA ($+1.4$) is ProxyCLIP H/14.
ProxyCLIP H/14 uses ViT-H as its CLIP backbone with ViT-B/14 DINOv2 as its internal proxy; FSA's QKV modulation is particularly effective here because it directly refines the high-resolution DINOv2-guided attention maps that this host constructs internally.
On the analogous B/16 configuration, TPA outperforms FSA ($+2.0$ vs.\ $+1.0$), and composing the two yields $+3.1$ (Table~5/composition table in the main paper), confirming that TPA adds orthogonal signal over FSA when the host is not at its absolute peak.
The H/14 exception is therefore a case where a per-image method is unusually well-matched to the specific host internals, not a systematic failure of TPA.

\paragraph{Robustness against Prototype Contamination.}
A concern with self-labelled bank construction is that the host's systematic errors could propagate into the prototypes: if the host confidently misclassifies a region, TPA would aggregate features from the wrong class.
Several factors mitigate this in practice.
First, the anchor criterion requires \emph{both} argmax consistency and softmax score exceeding $\tau = 2/C$: patches that pass this filter are those the host is most confident about, and systematic high-confidence errors on a diverse set of natural images are empirically rare.
Second, the running-mean aggregation averages features over potentially thousands of anchors across the adaptation pool; even if a small fraction are erroneous, their contribution to the final unit-norm prototype is diluted.
Third, the mean-centring step (Sec.~3.2 of the main paper) removes any per-pixel offset in the cosine scores, so a slightly imprecise prototype biases the auxiliary logit uniformly across classes rather than towards a specific wrong class.
The component ablation (Table~3 in the main paper) provides indirect empirical support: removing confidence-based anchor selection and replacing it with random anchor selection of the same count drops the average $\Delta$ from $+2.9$ to $+0.9$, confirming that anchor quality---not quantity---is the primary determinant of prototype usefulness.

\paragraph{Choice of Bank Extractor.}
The bank extractor ablation (Table~4 in the main paper) shows DINOv2 $>$ MAE $>$ SAM by a consistent margin.
The ordering reflects the degree to which each encoder's features are semantically organised at the patch level.
DINO's self-distillation objective encourages nearby patches of the same object to be mapped to nearby representations, producing tight per-class clusters that serve as effective class centroids~\cite{dino2021}.
MAE's masked reconstruction objective yields strong representations but without the explicit spatial grouping pressure;
SAM is trained for prompt-based segmentation and produces features organised around segment boundaries rather than semantic categories.
It is worth noting that TPA still yields positive gains with both MAE ($+1.8$ avg.) and SAM ($+1.3$ avg.), because even imperfect class-level aggregation provides more signal than no cross-image information at all.
The adoption of DINOv2 is therefore an empirical preference that could be revisited as new self-supervised encoders are released; the output-level interface of TPA imposes no constraint on the encoder choice.

\paragraph{Failure Modes and Degradation Cases.}
TPA can hurt on individual images in two scenarios.
The first is when the adaptation pool is too small to accumulate a reliable prototype for a low-frequency class: the prototype computed from only one or two anchor patches may reflect the specific pose or illumination of those anchors rather than the class distribution, pulling the inference-time cosine score in the wrong direction.
The $K_{\min}$ filter is designed to prevent this: if fewer than $K_{\min}=5$ anchor patches have been collected for class $c$, its prototype entry remains zero and the fusion reduces to the host's original log-probability.
The second scenario is distribution shift between the adaptation pool and the test image: if the pool images contain a class in canonical frontal poses while the test image shows an unusual viewpoint, the prototype may be a poor match.
The conservative weight $\alpha = 0.5$ limits the influence of the auxiliary logit to at most half the total score, preventing aggressive over-correction.
The overlap stress experiment (label-overlap robustness figure in the main paper) gives empirical evidence that TPA's degradation is bounded even under the adversarial low-overlap condition: per-class $\Delta$ stays non-negative on average, though with a narrower margin than under random or high overlap.

\section{Additional Qualitative Results and Failure Modes}
\label{sec:appendix-qualitative}

\begin{figure*}[t]
  \centering
  \includegraphics[width=\textwidth]{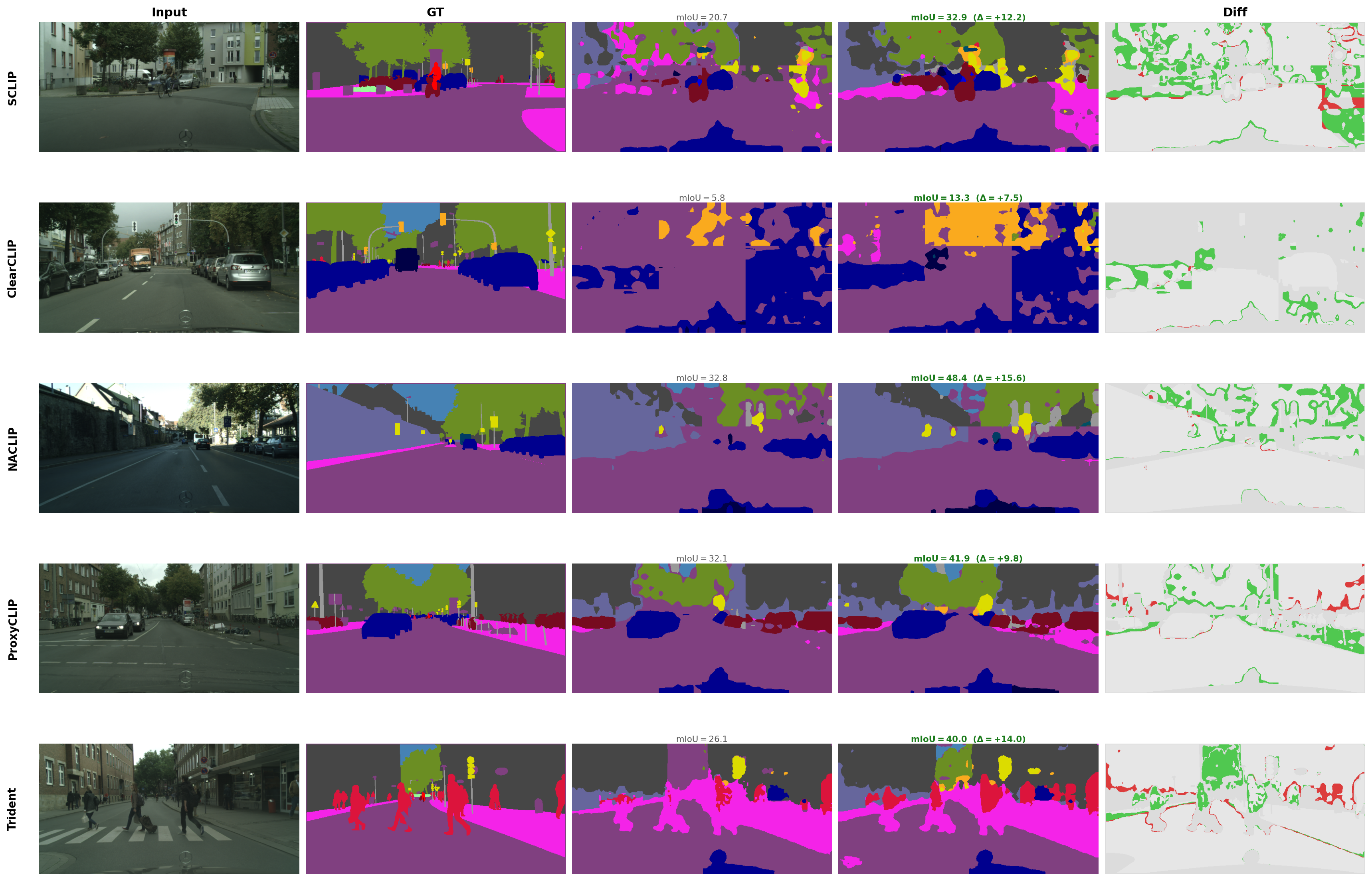}
  \caption{\textbf{Qualitative results on Cityscapes} (five hosts, best-gain image per host).}
  \label{fig:appendix-qual-city}
\end{figure*}

\begin{figure*}[t]
  \centering
  \includegraphics[width=\textwidth]{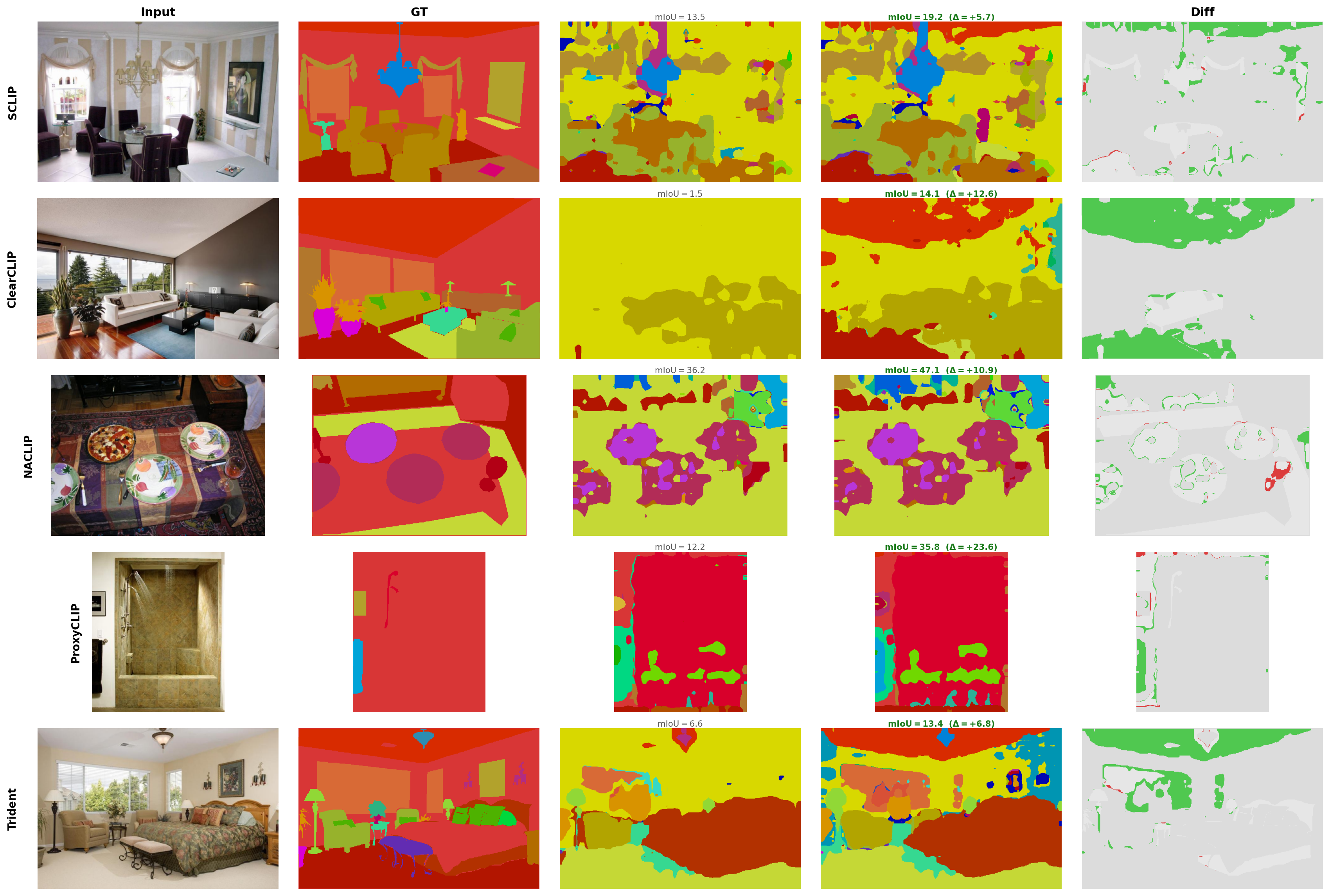}
  \caption{\textbf{Qualitative results on ADE20K} (five hosts, best-gain image per host).}
  \label{fig:appendix-qual-ade}
\end{figure*}

To provide a comprehensive understanding of TPA's behavior, Figures~\ref{fig:appendix-qual-ade}--\ref{fig:appendix-qual-city} extend the qualitative comparison in the main paper (VOC21) to three highly challenging and semantically dense benchmarks: ADE20K, COCO-Object, and Cityscapes. 

The same visualisation format is used: input image, ground truth, Base prediction, TPA-augmented prediction, and difference map (green: corrected by TPA; red: regressed; grey: both correct). Each row showcases the image with the highest overall per-image mIoU gain for each respective host model. As observed across all datasets, TPA exhibits a remarkable ability to unearth missing semantic regions and enforce crisp boundaries, even in cluttered indoor environments (ADE20K) or complex urban street scenes (Cityscapes), solely by leveraging the offline extracted DINO prototypes.

Crucially, the inclusion of the difference maps explicitly highlights not only the regions where TPA successfully recovers semantic layout (green), but also the occasional areas where TPA introduces local regressions (red). These red areas serve as excellent real-world failure mode visualizations. They typically occur at highly ambiguous object boundaries or when encountering out-of-distribution local textures that conflict with the accumulated class prototype. Openly presenting these localized regressions transparently illustrates the bounded degradation discussed in Sec.~\ref{sec:appendix-discussion}, confirming that while TPA overwhelmingly improves global dense prediction, its localized edits remain strictly dependent on the underlying feature separability of the frozen DINO space.

\begin{figure*}[t]
  \centering
  \includegraphics[width=\textwidth]{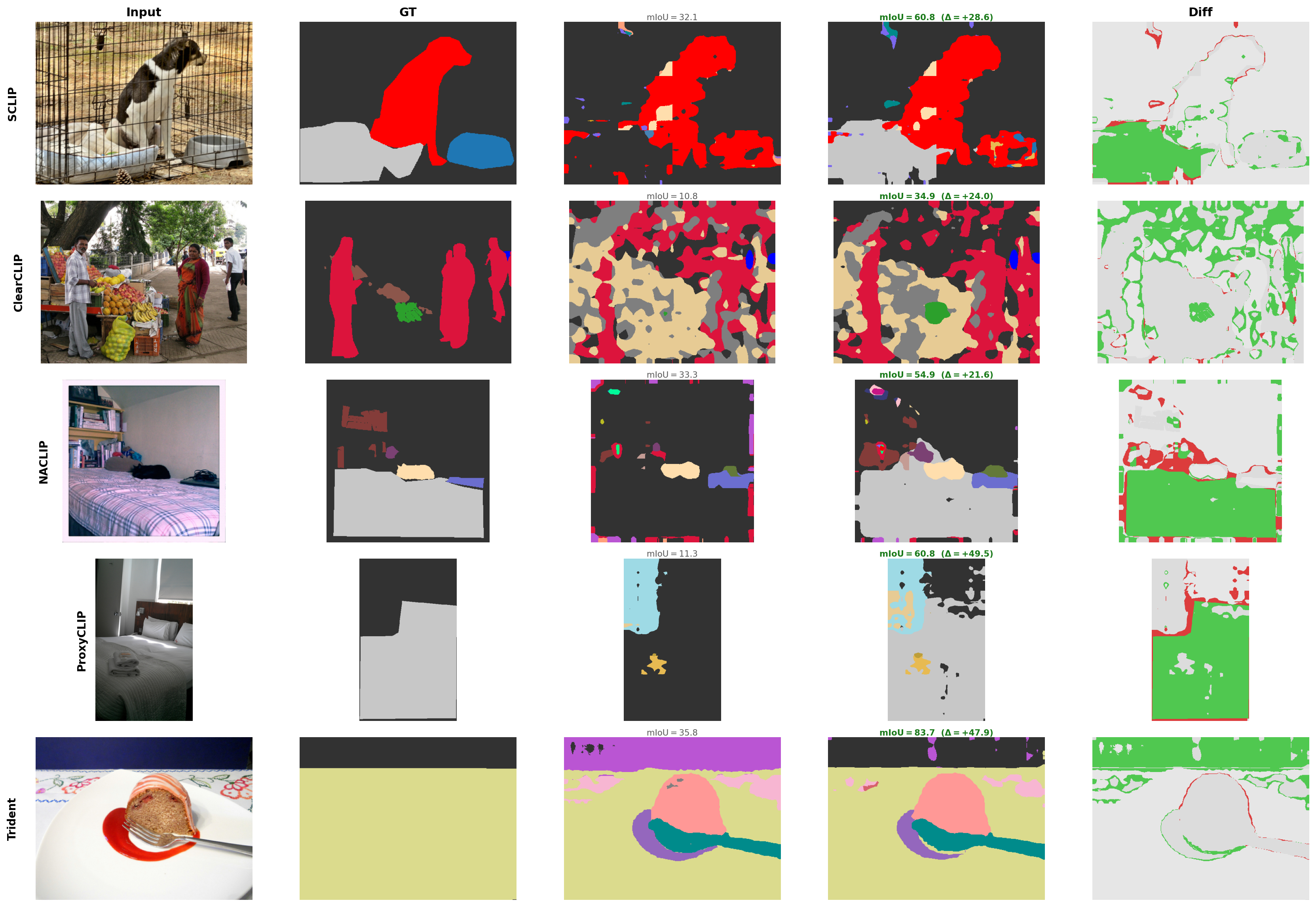}
  \caption{\textbf{Qualitative results on COCO-Object} (five hosts, best-gain image per host).}
  \label{fig:appendix-qual-coco-obj}
\end{figure*}

\clearpage

%% file: references.bib
@inproceedings{cheng2023pothole,
  title        = {{Pothole-YOLO}: A Single-Stage Instance Segmentation Method for Pothole Detection},
  author       = {Cheng, Jintao and Chen, Xingming and Chen, Weiwen and Huang, Zhuoxu and Wu, Jin and Fan, Rui and Tang, Xiaoyu},
  booktitle    = {International Conference on Autonomous Unmanned Systems},
  pages        = {450--465},
  year         = {2023},
  organization = {Springer}
}

@inproceedings{cheng2024mf,
  title        = {{MF-MOS}: A Motion-Focused Model for Moving Object Segmentation},
  author       = {Cheng, Jintao and Zeng, Kang and Huang, Zhuoxu and Tang, Xiaoyu and Wu, Jin and Zhang, Chengxi and Chen, Xieyuanli and Fan, Rui},
  booktitle    = {2024 IEEE International Conference on Robotics and Automation (ICRA)},
  pages        = {12499--12505},
  year         = {2024},
  organization = {IEEE}
}

@inproceedings{cheng2024mv,
  title        = {{MV-MOS}: Multi-View Feature Fusion for 3D Moving Object Segmentation},
  author       = {Cheng, Jintao and Chen, Xingming and Liang, Jinxin and Tang, Xiaoyu and Chen, Xieyuanli and Li, Dachuan},
  booktitle    = {2024 IEEE International Conference on Robotics and Biomimetics (ROBIO)},
  pages        = {7--13},
  year         = {2024},
  organization = {IEEE}
}

@article{cheng2025scale,
  title     = {Scale, Don't Fine-Tune: Guiding Multimodal {LLMs} for Efficient Visual Place Recognition at Test-Time},
  author    = {Cheng, Jintao and Li, Weibin and Luo, Jiehao and Tang, Xiaoyu and He, Zhijian and Wu, Jin and Zou, Yao and Zhang, Wei},
  journal   = {IFAC-PapersOnLine},
  volume    = {59},
  number    = {35},
  pages     = {97--102},
  year      = {2025},
  publisher = {Elsevier}
}

@inproceedings{clip2021,
  author    = {Radford, Alec and Kim, Jong Wook and Hallacy, Chris and others},
  title     = {Learning Transferable Visual Models From Natural Language Supervision},
  booktitle = {ICML},
  year      = {2021}
}

@misc{openclip2022,
  author = {Ilharco, Gabriel and Wortsman, Mitchell and Wightman, Ross and others},
  title  = {{OpenCLIP}},
  year   = {2022},
  note   = {Software release}
}

@inproceedings{dino2021,
  author    = {Caron, Mathilde and others},
  title     = {Emerging Properties in Self-Supervised Vision Transformers},
  booktitle = {ICCV},
  year      = {2021}
}

@article{dinov22023,
  author  = {Oquab, Maxime and others},
  title   = {{DINOv2}: Learning Robust Visual Features without Supervision},
  journal = {TMLR},
  year    = {2024}
}

@inproceedings{mae2022,
  author    = {He, Kaiming and others},
  title     = {Masked Autoencoders Are Scalable Vision Learners},
  booktitle = {CVPR},
  year      = {2022}
}

@inproceedings{sam2023,
  author    = {Kirillov, Alexander and others},
  title     = {Segment Anything},
  booktitle = {ICCV},
  year      = {2023}
}

@inproceedings{maskclip2022,
  author    = {Zhou, Chong and Loy, Chen Change and Dai, Bo},
  title     = {Extract Free Dense Labels from {CLIP}},
  booktitle = {ECCV},
  year      = {2022}
}

@inproceedings{sclip2024,
  author    = {Wang, Feng and others},
  title     = {{SCLIP}: Rethinking Self-Attention for Dense Vision-Language Inference},
  booktitle = {ECCV},
  year      = {2024}
}

@inproceedings{naclip2024,
  author    = {Hajimiri, Sina and others},
  title     = {Pay Attention to Your Neighbours: Training-Free Open-Vocabulary Semantic Segmentation},
  booktitle = {WACV},
  year      = {2025}
}

@inproceedings{proxyclip2024,
  author    = {Lan, Mengcheng and others},
  title     = {{ProxyCLIP}: Proxy Attention Improves {CLIP} for Open-Vocabulary Segmentation},
  booktitle = {ECCV},
  year      = {2024}
}

@inproceedings{clearclip2024,
  author    = {Lan, Mengcheng and others},
  title     = {{ClearCLIP}: Decomposing {CLIP} Representations for Dense Vision-Language Inference},
  booktitle = {ECCV},
  year      = {2024}
}

@inproceedings{gem2024,
  author    = {Bousselham, Walid and others},
  title     = {{GEM}: Grounding Everything --- Emerging Localization Properties in Vision-Language Transformers},
  booktitle = {CVPR},
  year      = {2024}
}

@inproceedings{trident2024,
  author    = {Shi, Yuheng and others},
  title     = {Harnessing Vision Foundation Models for High-Performance, Training-Free Open-Vocabulary Segmentation},
  booktitle = {CVPR},
  year      = {2025}
}

@inproceedings{t3a2021,
  author    = {Iwasawa, Yusuke and Matsuo, Yutaka},
  title     = {Test-Time Classifier Adjustment Module for Model-Agnostic Domain Generalization},
  booktitle = {NeurIPS},
  year      = {2021}
}

@inproceedings{tent2021,
  author    = {Wang, Dequan and others},
  title     = {{TENT}: Fully Test-Time Adaptation by Entropy Minimization},
  booktitle = {ICLR},
  year      = {2021}
}

@inproceedings{lame2022,
  author    = {Boudiaf, Malik and others},
  title     = {Parameter-Free Online Test-Time Adaptation},
  booktitle = {CVPR},
  year      = {2022}
}

@inproceedings{eata2022,
  author    = {Niu, Shuaicheng and others},
  title     = {Efficient Test-Time Model Adaptation without Forgetting},
  booktitle = {ICML},
  year      = {2022}
}

@inproceedings{cotta2022,
  author    = {Wang, Qin and others},
  title     = {Continual Test-Time Domain Adaptation},
  booktitle = {CVPR},
  year      = {2022}
}

@article{mlmp2025,
  author  = {Noori, Mehrdad and Osowiechi, David and Vargas Hakim, Gustavo Adolfo and
             Bahri, Ali and Yazdanpanah, Moslem and Dastani, Sahar and Beizaee, Farzad and
             Ben Ayed, Ismail and Desrosiers, Christian},
  title   = {Test-Time Adaptation of Vision--Language Models for
             Open-Vocabulary Semantic Segmentation},
  journal = {arXiv preprint arXiv:2505.21844},
  year    = {2025}
}

@inproceedings{fsa2025,
  author    = {Chi, Zhixiang and others},
  title     = {Plug-in Feedback Self-adaptive Attention in {CLIP} for Training-free
               Open-Vocabulary Segmentation},
  booktitle = {ICCV},
  year      = {2025}
}

@inproceedings{resclip2025,
  author    = {Yang, Yvhang and others},
  title     = {{ResCLIP}: Residual Attention for Training-free Dense Vision-language Inference},
  booktitle = {CVPR},
  year      = {2025}
}

@inproceedings{clipdinoiser2023,
  author    = {Wysocza{\'n}ska, Monika and Sim{\'e}oni, Oriane and Ramamonjisoa,
               Micha{\"e}l and Bursuc, Andrei and Trzci{\'n}ski, Tomasz and P{\'e}rez, Patrick},
  title     = {{CLIP-DINOiser}: Teaching {CLIP} a Few {DINO} Tricks for
               Open-Vocabulary Semantic Segmentation},
  booktitle = {ECCV},
  year      = {2024}
}

@article{pascalvoc2010,
  author  = {Everingham, M. and Van Gool, L. and Williams, C. K. I. and Winn, J. and Zisserman, A.},
  title   = {The {PASCAL} Visual Object Classes ({VOC}) Challenge},
  journal = {IJCV},
  year    = {2010}
}

@inproceedings{pcontext2014,
  author    = {Mottaghi, Roozbeh and others},
  title     = {The Role of Context for Object Detection and Semantic Segmentation in the Wild},
  booktitle = {CVPR},
  year      = {2014}
}

@article{ade20k2019,
  author  = {Zhou, Bolei and others},
  title   = {Semantic Understanding of Scenes Through the {ADE20K} Dataset},
  journal = {IJCV},
  year    = {2019}
}

@inproceedings{coco2014,
  author    = {Lin, Tsung-Yi and others},
  title     = {{Microsoft COCO}: Common Objects in Context},
  booktitle = {ECCV},
  year      = {2014}
}

@inproceedings{cocostuff2018,
  author    = {Caesar, Holger and Uijlings, Jasper and Ferrari, Vittorio},
  title     = {{COCO-Stuff}: Thing and Stuff Classes in Context},
  booktitle = {CVPR},
  year      = {2018}
}

@inproceedings{cityscapes2016,
  author    = {Cordts, Marius and others},
  title     = {The {Cityscapes} Dataset for Semantic Urban Scene Understanding},
  booktitle = {CVPR},
  year      = {2016}
}
